\pdfoutput=1
\documentclass{article}

\PassOptionsToPackage{numbers, compress, sort}{natbib}

\usepackage[preprint]{neurips_2023}

\usepackage[utf8]{inputenc} 
\usepackage[T1]{fontenc}    
\usepackage{hyperref}       
\usepackage{url}            
\usepackage{booktabs}       
\usepackage{amsfonts}       
\usepackage{nicefrac}       
\usepackage{microtype}      
\usepackage{xcolor}         
\usepackage{graphicx}
\usepackage{multirow}
\usepackage{multicol}
\usepackage{subcaption}
\usepackage{amsmath}

\usepackage{capt-of}
\usepackage{subcaption}
 \usepackage{setspace}

\hypersetup{
    colorlinks = true,
    linkbordercolor = {white},
    citecolor=blue
}

\newcommand{\CUT}[1]{}

\newcommand{\chenyang}[1]{{\color{red}{[chenyang: #1]}}}

\newcommand{\ie}{\textit{i}.\textit{e}.}
\newcommand{\eg}{\textit{e}.\textit{g}.}
\newcommand{\etc}{\textit{etc}.}

\title{Inserting Anybody in Diffusion Models via Celeb Basis}

\author{
    Ge Yuan\textsuperscript{\rm 1,2}\quad Xiaodong Cun\textsuperscript{\rm 2}\quad  Yong Zhang\textsuperscript{\rm 2} \quad Maomao Li\textsuperscript{\rm 2}\thanks{Corresponding authors} \quad  Chenyang Qi\textsuperscript{\rm 2,3}\\ 
    \textbf{Xintao Wang\textsuperscript{\rm 2}\quad  Ying Shan\textsuperscript{\rm 2} \quad Huicheng Zheng\textsuperscript{\rm 1}\footnotemark[1]}\\ \\
    \textsuperscript{\rm 1}~Sun Yat-sen University \quad \textsuperscript{\rm 2}~Tencent AI Lab \quad \textsuperscript{\rm 3}~HKUST \\ \\
    \url{http://celeb-basis.github.io}
}

\begin{document}

\maketitle

\begin{figure}[h]
    \vspace{-2em}
    \centering
    \includegraphics[width=\textwidth]{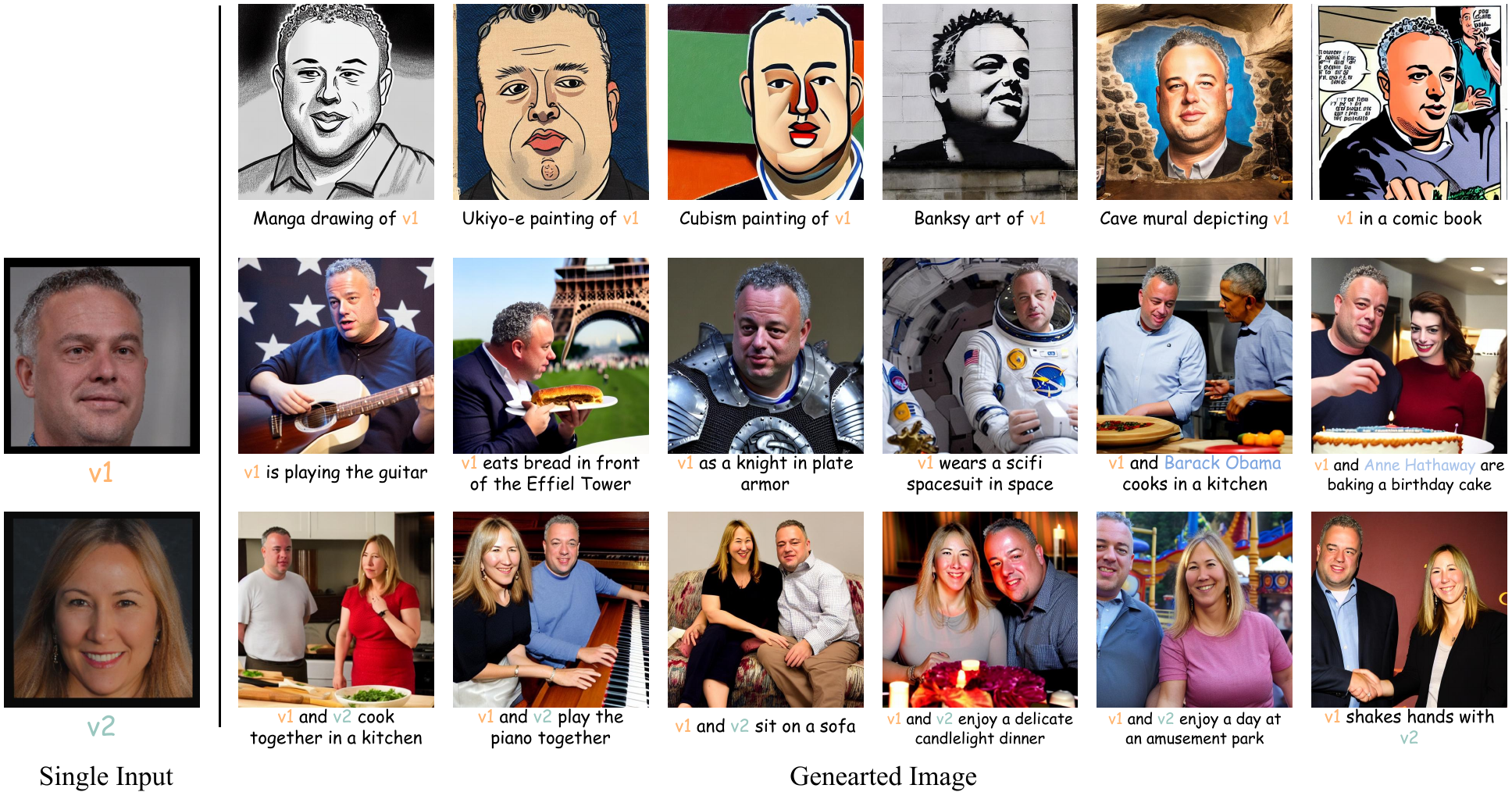}
    \vspace{-1em}
    \caption{Given a single facial photo~(\textcolor{orange}{$v1$} or \textcolor{teal}{$v2$}) as a tunable sample, the proposed method can insert this identity into the trained text-to-image model, \eg, Stable Diffusion~\cite{stable-diffusion}, where the new person\textcolor{orange}{~($v1$)} can act like the original concept in the trained model and interact with another newly trained concept\textcolor{teal}{~($v2$)}. Note that the input images are randomly generated from StyleGAN~\cite{stylegan}.}
    \label{fig:teaser}
\end{figure}

\begin{abstract}
Exquisite demand exists for customizing the pretrained large text-to-image model, \eg, Stable Diffusion, to generate innovative concepts, such as the users themselves. However, the newly-added concept from previous customization methods often shows weaker combination abilities than the original ones even given several images during training. 
We thus propose a new personalization method that allows for the seamless integration of a unique individual into the pre-trained diffusion model using just \textit{one facial photograph} and only \textit{1024 learnable parameters} under \textit{3 minutes}. So as we can effortlessly generate stunning images of this person in any pose or position, interacting with anyone and doing anything imaginable from text prompts. 
To achieve this, we first analyze and build a well-defined celeb basis from the embedding space of the pre-trained large text encoder. Then, given one facial photo as the target identity, we generate its own embedding by optimizing the weight of this basis and locking all other parameters.
Empowered by the proposed celeb basis, the new identity in our customized model showcases a better concept combination ability than previous personalization methods. Besides, our model can also learn several new identities at once and interact with each other where the previous customization model fails to. The code will be released.
\end{abstract}

\section{Introduction}
\label{sec:intro}


Vast image-text pairs~\cite{laion} during training and the powerful language encoder~\cite{clip} enable the text-to-image models~\cite{stable-diffusion, imagen, dalle2} to generate diverse and fantastic images from simple text prompts. Though the generated images are exquisite, they may still fail to satisfy the users' demands since some concepts are not easy to be described by the text prompt~\cite{textual-inversion}. For instance, individuals frequently post self-portraits on social media platforms, where the pre-trained text-to-image models struggle to produce satisfactory pictures of them despite receiving comprehensive instructions. This shortcoming makes these models less attractive to general users. 

Recent works~\cite{custom-diffusion, dreamartist,dreambooth,textual-inversion} solve this problem via efficiently tuning the parameters of the model for personalization usage. For example, 
these techniques insert the new concepts~(\eg, a specific bag, a dog, or a person) into the model by representing them as rarely-used pseudo-words~\cite{dreambooth}~(or text-embeddings~\cite{dreamartist,textual-inversion,custom-diffusion, fast-encoder})
and finetuning the text-to-image model with a few of these samples~\cite{dreambooth, textual-inversion}. After training, these pseudo-words or embeddings can be used to represent the desired concept and can also perform some combination abilities.
However, these methods often struggle to generate the text description-aligned image with the same concept class~(\eg, the person identities)~\cite{custom-diffusion, dreambooth}. For instance, the original Stable Diffusion~\cite{stable-diffusion} model can successfully generate the image of the text prompt:`` \textit{Barack Obama and Anne Hathaway are shaking hands.}" \CUT{but}
Nevertheless, in terms of generating two newly-learned individuals, the previous personalized methods~\cite{custom-diffusion,dreambooth,textual-inversion} fall short of producing the desired identities as depicted in our experiments (Figure~\ref{fig:exp_action}).

In this work, we focus on injecting the most specific and widely-existing concept, \ie, the human being, into the diffusion model seamlessly. 
Drawing inspiration from the remarkable 3DMM~\cite{3dmm}, which ingeniously represents novel faces through a combination of mean and weight values derived from a clearly defined basis, we build a similar basis to the embeddings of the celebrity names in pretrained Stable Diffusion. In this way, we are capable of representing any new person in the trained diffusion model via the basis coefficients. 



We first collect a bunch of celebrity names from the Internet and filter them by the pre-trained text-to-image models~\cite{stable-diffusion}.
By doing so, we obtain 691 well-known names and extract the text embedding by the tokenizer of the CLIP~\cite{clip}. Then, we construct a \textit{celeb basis} via Principal Component Analysis~(PCA~\cite{pca}). To represent a new person with PCA coefficients, we use a pre-trained face recognition network~\cite{arcface} as the feature extractor of the given photo and learn a series of coefficients to re-weight the celeb basis, so that the new face can be recognized by the pre-trained CLIP transformer encoder. During the process, we only use a single facial photo and fix the denoising UNet and the text encoder of Stable Diffusion to avoid overfitting. After training, we only need to store the 1024 coefficients of the celeb basis to represent the newly-added identity since the basis is shared across the model. Yet simple, the concept composition abilities~\cite{stable-diffusion} of the trained new individual is well-preserved, as we only reweight the text embeddings of the trained CLIP model and freeze the weights in the diffusion process.
Remarkably, the proposed method has the ability to produce a strikingly realistic photo of the injected face in any given location and pose. Moreover, it opens up some new possibilities such as learning multiple new individuals simultaneously and facilitating seamless interaction between these newly generated identities.

The contributions of the paper are listed as follows:

\begin{itemize}
    \item We propose celeb basis, a basis built from the text embedding space of the celebrities' names in the text-to-image model and verify its abilities, such as interpolation.
    \item Based on the proposed celeb basis, we design a new personalization method for the text-to-image model, which can remember any new person from a single facial photo using only 1024 learnable coefficients.
    \item Extensive experiments show our personalized method has more stable concept composition abilities than previous works, including generating better identity-preserved images and interacting with new concepts.
\end{itemize}


\section{Related Work}
\vspace{-1.em}
\label{sec:related}
\textbf{Image Generation and Editing.}
Given a huge number of images as the training set, deep generative models target to model the distribution of training data and synthesize new realistic images through sampling. Various techniques have been widely explored, including GAN~\cite{gan,stylegan}, VAE~\cite{vae,vqvae}, Autoregressive~\cite{dalle2,taming,VQVAE2,dalle}, flow~\cite{kingma2018glow,dinh2014nice}.
Recently, diffusion models~\cite{ddpm, ddim} gain increasing popularity for their stronger abilities of text-to-image generation~\cite{imagen,dalle2,stable-diffusion}. 
Conditioned on the text embedding of pre-trained large language models~\cite{clip}, these diffusion models are iterative optimized using a simple denoising loss.
During inference, a new image can be generated from sampled Gaussian noise and a text prompt.
Although these diffusion models can synthesize high-fidelity images, they have difficulties in generating less common concepts~\cite{chen2022reimagen} or controlling the identity of generated objects~\cite{textual-inversion}. 
Current editing methods are still hard to solve this problem, \eg, directly blending the latent of objects~\cite{blended,blended_latent} to the generated background will show the obvious artifacts and is difficult to understand the scenes correctly~\cite{paint-by-example}. On the other hand, attention-based editing works~\cite{null, pix2pix-zero, qi2023fatezero, p2p} only change the appearance or motion of local objects, which can not generate diverse new images with the same concept~(\eg, human and animal identity) from the referred image.


\textbf{Model Personalization.}
Different from text-driven image editing, tuning the model for the specific unseen concept, \ie, personalized model, remembers the new concepts of the reference images and can synthesize totally unseen images of them, \eg, appearance in a new environment, interaction with other concepts in the original stable diffusion.
For generative adversarial neworks~\cite{stylegan,karras2020analyzing}, personalization through GAN inversion has been extensively studied. This progress typically involves finetuning of the generator~\cite{nitzan2022mystyle, roich2022pivotal}, test-time optimization of the latents~\cite{abdal2019image2stylegan}, or a pre-trained encoder~\cite{richardson2021encoding}. Given the recent diffusion generative model~\cite{stable-diffusion, imagen}, it is straightforward to adopt previous GAN inversion techniques for the personalization of diffusion models.
Dreambooth~\cite{dreambooth} finetunes all weight of the diffusion model on a set of images with the same identity and marks it as the specific token. 
Meanwhile, another line of works ~\cite{textual-inversion, dreamartist, kawar2022imagic} optimizes the text embedding of special tokens (\eg, $V^*$) to map the input image while freezing the diffusion model.
Later on, several works~\cite{custom-diffusion, continualdiffusion, perfusion} combine these two strategies for multi-concept interaction and efficient finetuning taking less storage and time. These methods focus on general concepts in the open domain while struggling to generate interactions between fine-grained concepts, \ie human beings with specific identities.
Since most of the previous works require the tuning in the test time, training inversion encoders are also proposed to generate textual embedding from a single image in the open domain~(\eg, UMM~\cite{umm}, ELITE~\cite{wei2023elite}, and SuTI~\cite{suTI}), or in human and animal domain~(\eg, Taming-Encoder~\cite{jia2023taming}, Instant-Booth~\cite{shi2023instantbooth}, E4T~\cite{fast-encoder}). However, a general human identity-oriented embedding is difficult to be obtained from a naively optimized encoder, and tuning the Stable Diffusion on larger-scale images often causes the concept forget. In contrast, our method focuses on a better representation of identity embedding in the diffusion model~(celeb basis in Sec.~\ref{sec:celeb_basis}), which significantly eases the process of optimization such that we only need 1024 parameters to represent an identity more correctly as in
Sec.~\ref{sec:finetune_model} and stronger concept combination abilities.

\textbf{Identity Basis.}
Representing the human identity via basis is not new in traditional computer vision tasks. \eg, in human face modeling, 3D Morphable Models~\cite{3dmm} and its following models~\cite{flame, Yang_2020_CVPR} scans several humans and represent the shape, expression, identity, and pose as the PCA coefficients~\cite{pca}, so that the new person can be modeled or optimized via the coefficients of the face basis. Similar ideas are also used for face recognition~\cite{kim2002face}, where the faces in the dataset are collected and built on the basis of the decision bound. Inspired by these methods, our approach takes advantage of the learned celebrity names in the pre-trained text-to-image diffusion model, where we build a basis on this celebrity space and generate the new person via a series of learned coefficients.




\section{Method}
\label{sec:method}
\vspace{-1em}


Our method aims to introduce a new identity to the pre-trained text-to-image model, \ie, Stable Diffusion~\cite{stable-diffusion}, from a single photo via the optimized coefficients of our self-built celeb basis. 
So that it can memorize this identity and generate new images of this person in any new pose and interact with other identities via text prompts. 
To achieve this, we first analyze and build a celeb basis on the embedding space of the text encoder through the names of the celebrities~(Sec.~\ref{sec:celeb_basis}). 
Then, we design a face encoder-based method to optimize the coefficients of the celeb basis for text-to-image diffusion model customization~(Sec.~\ref{sec:finetune_model}).

\if
With the help of Large Language Models (LLMs)~\cite{clip} and pretrained on large paired datasets~\cite{laion}, text-to-image diffusion models~\cite{stable-diffusion, dalle2, imagen} can naturally generate the images of some celebs through indicating the names.
Providing the prompts about the celeb names, the LLMs encode the texts into the latent codes as a condition to control the synthesis process of diffusion models.
In this way, users can drive the action of the target celebs, e.g. "playing the guitar", "riding a motorbike", "talks with each other".
However, when providing the names of a person that is not recognized, such as a name of your friend, the text-to-image synthesizing models can hardly generate the identity-consistent images.
To solve this, previous methods personalize a new concept by finetuning the model weights~\cite{} or learning word embeddings~\cite{} to adapt the model.
Although these methods achieve good performance on some object concepts, they cannot handle the new human identity and control the interaction like the recognized celebs, especially when given only one face image, as shown in Figure~\ref{}.

The key reason for this is the domain gap between the ground-truth celeb embedding space and the learned one.~\chenyang{better to give an experiment evidence e.g., analyze their mean, variance, entropy}
After training on the images of a new person, the learned representations may lie in a space which is far from the real person space.
Directly making the model converged on input images without a specific direction leads to the under-fitting or concept-forgetting problems.
We propose to fit the target human into the person space using the celeb basis.
Using a linear combination of celeb embeddings to represent a new person, the model can realize that the obtained representation is a `person' and generate precise results consistent with the prompt.
\fi

\vspace{-1em}
\subsection{Celeb Basis} 
\label{sec:celeb_basis}
\vspace{-1em}

\begin{figure}[t]
    \centering
    \vspace{-2em}
    \includegraphics[width=1\textwidth]{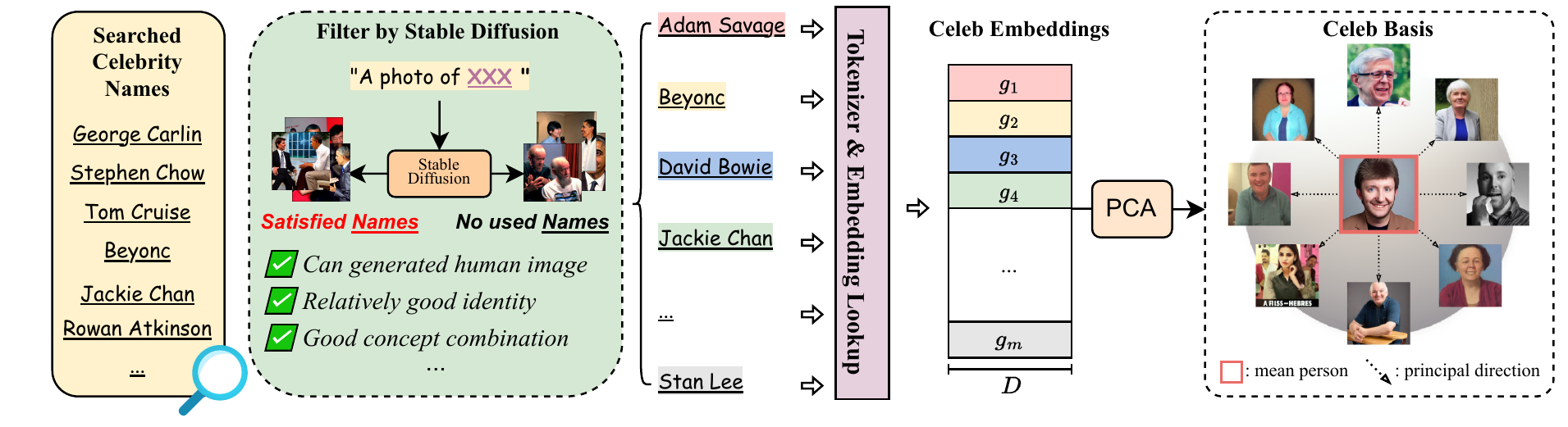}
    \caption{The building process of the proposed Celeb Basis.
    First, we collect about 1,500 celebrity names as the initial collection. Then, we manually filter the initial one to $m=691$ names, based on the synthesis quality of text-to-image diffusion model~\cite{stable-diffusion} with corresponding name prompt. 
    Later, each filtered name is tokenized and encoded into a celeb embedding group $g_i$. Finally, we conduct Principle Component Analysis to build a compact orthogonal basis, which is visualized on the right.}
    \label{fig:method_celeb_basis}
    \vspace{-1em}
\end{figure}





\textbf{Preliminary: Text Embeddings in Text-to-Image Diffusion Models.}
In the text-to-image model, given any text prompts $u$, the tokenizer of typical text encoder model $e_{\rm{text}}$, \eg, BERT~\cite{bert} and CLIP~\cite{clip}, divides and encodes $u$ into $l$ integer tokens by order.
Correspondingly, by looking up the dictionary, an embedding group $g =[v_{1},...,v_{l}]$ consisting of $l$ word embeddings can be obtained, where each embedding $v_{i}\in\mathbb{R}^{d}$.
Then the text transformer $\tau_{\rm{text}}$ in $e_{\rm{text}}$ encodes $g$  and generates text condition $\tau_{\rm{text}}(g)$.
The condition $\tau_{\rm{text}}(g)$ is fed to the conditional denoising diffusion model $\epsilon_\theta(z_t,t,\tau_{\rm{text}}(g))$ and synthesize the output image following an iterative denoising process~\cite{ddpm}, where $t$ is the timestamp, $z_t$ is an image or latent noised to $t$. Previous text-to-image model personalization methods~\cite{textual-inversion,dreambooth,dreamartist} have shown the importance of text embedding $g$ in personalizing semantic concepts. However, in text-to-image models' personalization, they only consider it as an optimization goal~\cite{dreamartist,textual-inversion,custom-diffusion}, instead of improving its representation.

\begin{figure}[h]
    \centering
    \includegraphics[width=\textwidth]{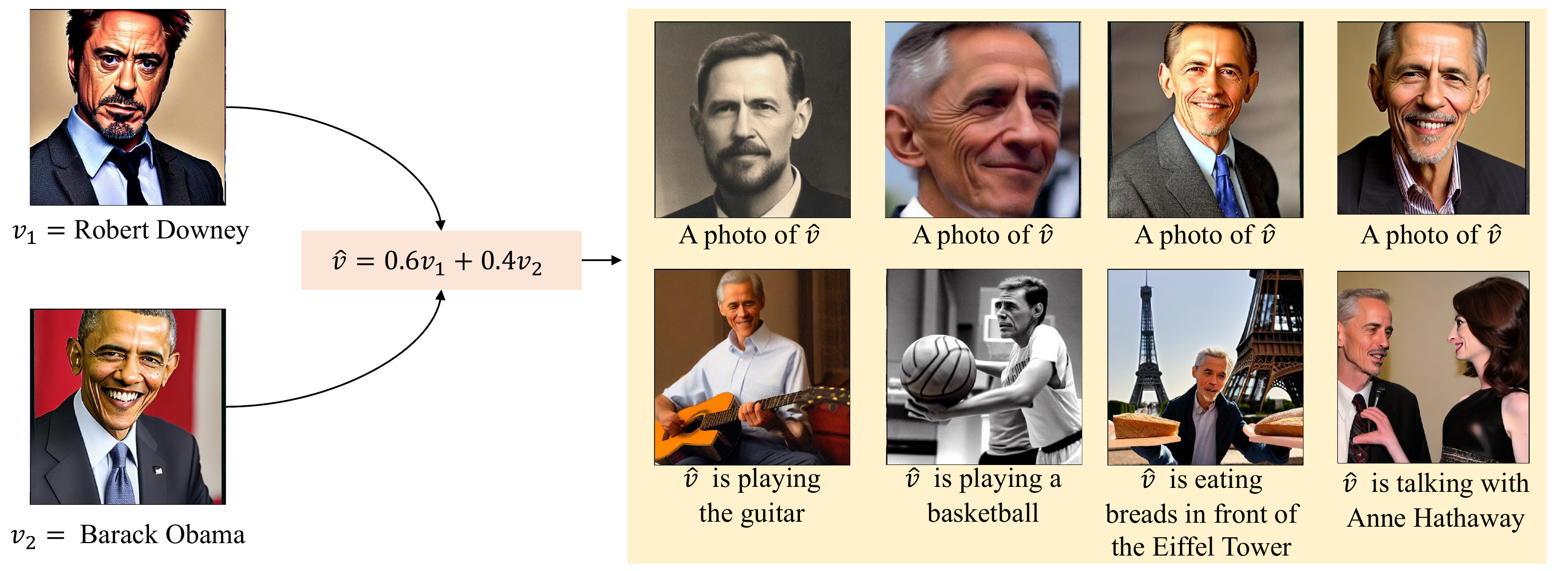}
    \vspace{-1.5em}
    \caption{The interpolated text-embedding of two celebrities is still a human~(top row) and it also can perform strong\CUT{er} concept combination abilities in the pretrained Stable Diffusion~\cite{stable-diffusion}~(bottom row). }
    \label{fig:interpolation}
\end{figure}

\textbf{Interpolating Abilities of Text Embeddings.}
Previous works have shown that text embedding mixups~\cite{guo2020nonlinear} benefit text classification. To verify the interpolation abilities in text-to-image generation, we randomly pick two celebrity names embeddings $v_1$ and $v_2$, 
and linearly combine them as $\hat{v}=\lambda v_1 + (1-\lambda) v_2$, where $0<\lambda<1$. Interestingly, the generated image of the interpolated embedding $\hat{v}$ also contains a human face as shown in Figure~\ref{fig:interpolation}, and all the generated images perform well in acting and interacting with other celebrities. 
Motivated by the above finding, we build a celeb basis so that each new identity can lie in the space formed by celebrity embeddings.

\textbf{Build Celeb Basis from Embeddings of the Collected Celebrities.}
As shown in Figure~\ref{fig:method_celeb_basis}, 
we first crawl about 1,500 celebrity names from Wikipedia as the initial collection.
Then, we build a manual filter based on the trained text-to-image diffusion model~\cite{stable-diffusion} by constructing the prompts of each name and synthesizing images.
A satisfied celeb name should have the ability to generate human images with prompt-consistent identity and interact with other celebs in synthesized results. Overall, we get $m=691$ satisfied celeb names where each name $u_i, i\in\{1,..., m\}$ can be tokenized and encoded into a celeb embedding group $g_i = [v^i_{1}, ..., v^i_{k_i}]$, notice that the length $k_i$ of each celeb embedding group $g_i$ might not the same since each name may contain multiple words~(or the tokenizer will be divided the word by sub-words). To simplify the formula, we compose the nonrepetitive embeddings so that each $g_i$ only contains the first two embeddings~(\ie, $k_i=2$ for all $m$ celebrities).
Using $C_1$ and $C_2$, \ie, $C_k = [v^1_k,...,v^m_k]$, to denote the first and second embeddings of each $g_i$ respectively,  we can roughly understood them as the \textit{first name and last name embedding sets.}
To further build a compact search space, inspired by 3DMM~\cite{3dmm} which uses PCA~\cite{pca} to map high-dimensional scanned 3D face coordinates into a compact lower-dimensional space, for each embedding set $C_k$, 
we calculate its mean $\overline{C}_k = \frac{1}{m}\sum\nolimits^{m}_{i=1}v^i_{k} $ and PCA mapping matrix $ B_{k} = {\rm{PCA}}(C_k, p) $,
where $\overline{C}_k\in\mathbb{R}^{d}$ and 
${\rm{PCA}}(X,p)$ indicates the PCA operation that reduces the second dimension of matrix $X\in\mathbb{R}^{m\times d}$ into $p$ $(p<d)$ principal components, \ie, 
$B_{k}=[b^1_k,...,b^p_k]$.
As shown in Fig~\ref{fig:method_celeb_basis}, the mean embedding $\overline{C}_k$ still represents a face and we can get the new face via some coefficients applied to $B_{k}$.

Overall, our celeb basis is defined on two basis $[\overline{C}_1, B_1]$ and $[\overline{C}_2, B_2]$ working like the first and last name. We use the corresponding principle components $A_1$ and $A_2$~(where $A_k= [\alpha^1_k, ..., \alpha^p_k] $) to represent new identities. Formally, for each new person $\hat{g}$, we use two $p$-dimensional coefficients of the celeb basis and can be written by:
\begin{align}
\label{eq:embedding_group}
        \hat{g} = [\hat{v}_1, \hat{v}_2], \quad
        \hat{v}_k = \overline{C}_k + \sum_{x=1}^p \alpha^x_k b^x_k,
\end{align}
In practice, $p$ equals 512 as discussed in the ablation experiments. 

To control the generated identities, we optimize the coefficients with the help of a face encoder as the personalization method. We introduce it in the below section.




\subsection{Stable Diffusion Personalization via Celeb Basis}
\label{sec:finetune_model}

\begin{figure}[t]
    \centering
    \includegraphics[width=\textwidth]{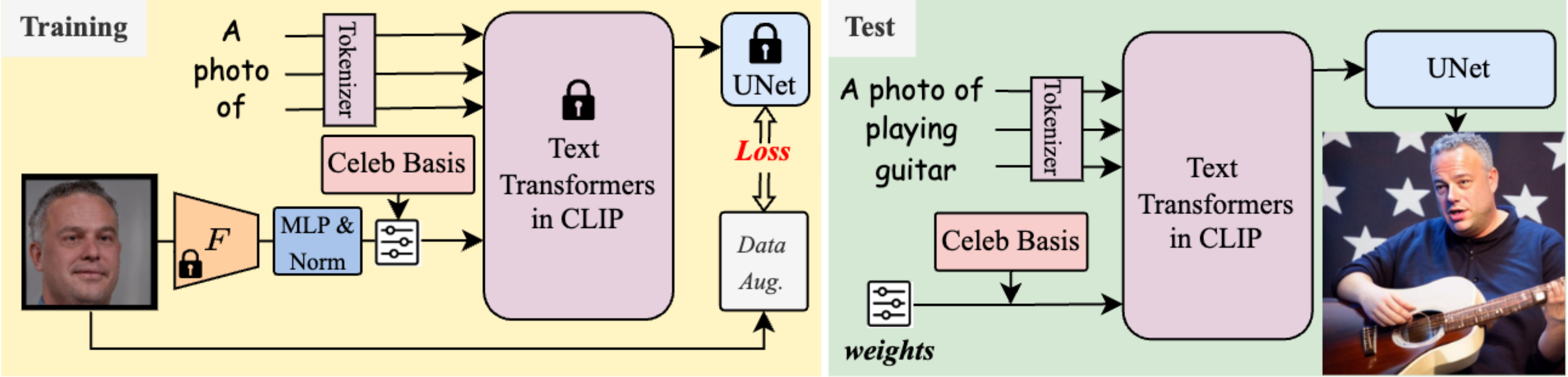}
    \caption{During training~(left), we optimize the coefficients of the celeb basis with the help of a fixed face encoder. During inference~(right), we combine the learned personalized weights and shared celeb basis to generate images with the input identity.}
    \label{fig:personalization}
\end{figure}

\textbf{Fast Coefficients Optimization for Specific Identity.} 
Given a single facial photo, we use the proposed celeb basis to embed the given face image $x$ of the target identity into the pretrained text-to-image diffusion model as shown in Fig.~\ref{fig:personalization}.
Since direct optimization is hard to find the optimized weight, we consider using the pre-trained state-of-the-art face recognition models $F$, \ie, ArcFace~\cite{arcface}, to capture the identity-discriminative patterns. In detail, we adopt the $F$ to extract 512 dimension face embedding as priors.
Then a single-layer MLP followed by an $L_2$-normalization is used to map the face priors into the modulating coefficients $A_1$ and $A_2$.
Following the Eq.~\ref{eq:embedding_group}, we can obtain the embedding group $\hat{g}$ of the $x$ using the pre-defined basis.
By representing the text prompt of $\hat{g}$ as $V^*$, we can involve $V^*$ to build the training pairs between the text prompt of input face and ``\textit{A photo of $V^*$}'', ``\textit{A depiction of $V^*$}'', \etc. Similar to previous works~\cite{dreambooth, textual-inversion, custom-diffusion}, we only use simple diffusion denoising loss~\cite{ddpm}:
\begin{align}
    \mathbb{E}_{\epsilon\sim N(0,1),x,t,g}[\Vert \epsilon - \epsilon_{\theta}(z_t,t,\tau_{\rm{text}}(g)) \Vert],
\end{align}
where $\epsilon$ is the unscaled noise sample, $g$ denotes the text embeddings containing $\hat{g}$.
During training, only the weights of MLP need to be optimized, while other modules,  including the celeb basis, face encoder $F$, CLIP transformer $\tau_{\rm{text}}$, and UNet $\epsilon_\theta$ are fixed. Thus, the original composition abilities of the trained text-to-image network are well-preserved, avoiding the forgetting problem.
Since we only have a single photo, we use color jitter, random resize, and random shift as data augmentations on the supervision to avoid overfitting. Notice that, we find our augmentation method can work well even though there are no regularization datasets which is important in previous methods~\cite{dreambooth,textual-inversion,custom-diffusion}, showing the strong learning abilities of the proposed methods. Since the proposed methods only involve a few parameters, it only takes almost \textit{3 minutes} for each individual on an NVIDIA A100 GPU, which is also much faster than the previous. 

\textbf{Testing.}
After training, only two groups of coefficients $A_1$ and $A_2$ applied to the principal celeb basis components need to be saved. In practice, the number of principal components of each group is $p=512$, coming to only \textit{1024 parameters} and \textit{2-3KB} storage consumption for half-precision floatings.
Then, users can build the prompt with multiple action description prompts~(\eg ``\textit{A photo of $V^*$ is playing guitar}'') to synthesize the satisfied images as described in Figure~\ref{fig:personalization}.



\textbf{Multiple Identities Jointly Optimization.}
Most previous methods only work on a single new concept~\cite{dreamartist,dreambooth}, Custom Diffusion~\cite{custom-diffusion} claim their method can generate the images of multiple new concepts~(\eg, the sofa and cat). However, for similar concepts, \eg, the different person, their method might not work well as in our experiments. 
Besides, their method is still struggling to learn multiple~($>3$) concepts altogether as in their limitation. 
Differently, we can learn multiple identities~(\eg, 10) at once using a shared MLP mapping layer as in Fig~\ref{fig:personalization}. In detail, we simply extend our training images to 10 and jointly train these images using a similar process as single identities. Without a specific design, the proposed method can successfully generate the weight of each identity. After training, our method can perform interactions between each new identity while the previous methods fail to as shown in the experiments.
More implementation details are in the supplementary.


\begin{figure}[t!]
    \centering
    \includegraphics[width=0.98\textwidth]{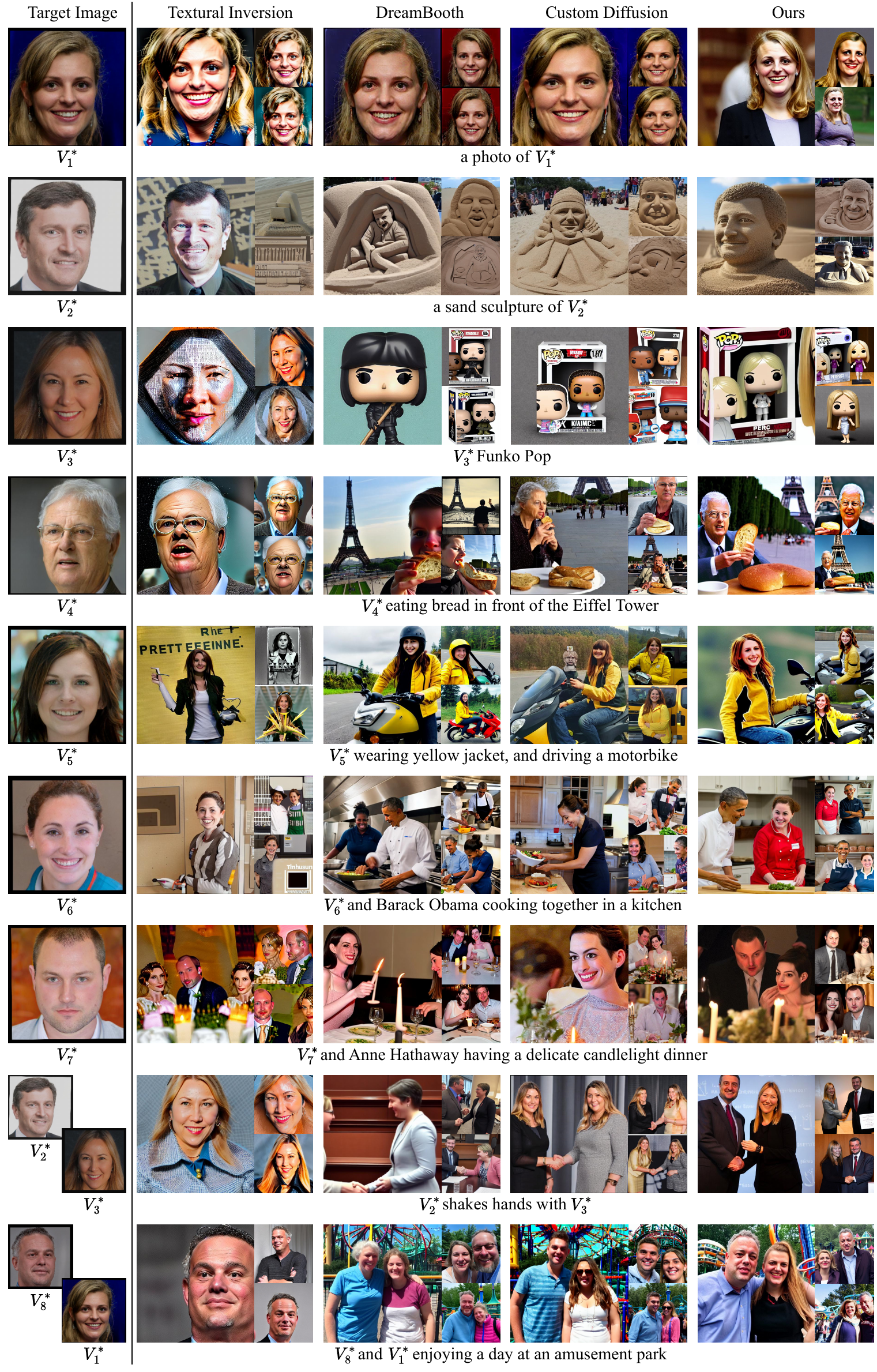}
    \vspace{-0.8em}
    \caption{We compare several different abilities between our method and baselines~(Textural Inversion~\cite{textual-inversion}, Dreambooth~\cite{dreambooth}, and Custom Diffusion~\cite{custom-diffusion}). }
    \label{fig:exp_action}
\end{figure}

\section{Experiments}
\vspace{-1em}




\subsection{Datasets and Metrics}

\textbf{Datasets.}
We conduct experiments on the self-collected 2k synthetic facial images generated by StyleGAN~\cite{stylegan}. By utilizing synthetic faces for assessing the effectiveness of generative models, one can ease the reliance on the dataset foundation of the initial pre-trained text-to-image model. We also perform some experiments on the real photo of the individual, more results and comparisons are shown in the supplemental materials.


\textbf{Metrics.} 
First, we assess the performance of the generated images by utilizing objective metrics. For instance, we calculate the consistency between the prompt and generated image through CLIP score~\cite{clip}, which is denoted as ``\textbf{Prompt}" in tables. Additionally, ensuring identity consistency and clarity of facial features are crucial aspects of our task. Therefore, we evaluate identity similarity using a pretrained face recognition encoder~\cite{arcface} and mark it as ``\textbf{Identity}". Furthermore, to demonstrate the rationality behind generation, we also calculate the rate of successful face detection~(``\textbf{Detect}") via a pretrained face detector~\cite{arcface}. Lastly, user studies are conducted to evaluate text-image alignment along with identity and photo qualities.


\begin{table}[b]
        \vspace{-2em}
      \caption{Quantitative comparisons.}
      \label{tab:sota}
      \centering
      \resizebox{\textwidth}{!}{
    \begin{tabular}{llllccccc}
    \toprule
    \multirow{2}{*}{Methods} & \multicolumn{3}{c}{Objective Metrics$\uparrow$} & \multicolumn{3}{c}{User Study$\uparrow$} & \multirow{2}{*}{\#Params$\downarrow$} & \multirow{2}{*}{Time$\downarrow$} \\ 
    \cmidrule(r){2-4} \cmidrule(r){5-7} & Prompt & Identity & Detect & Quality & Text & Identity &  & (min) \\
    \midrule
    Textual Inversion~\cite{textual-inversion} & 0.1635 & \textbf{0.2958} & \textbf{92.86\%} & 2.23 & 1.88 & 2.55 & 1,536 & 24 \\
    Dreambooth~\cite{dreambooth} & 0.2002 & 0.0512 & 54.76\% & 3.32 & 3.70 & 2.75 & $9.83\times10^8$ & 16 \\
    Custom Diffusion~\cite{custom-diffusion} & \textbf{0.2608} & 0.1385 & 80.39\% & 3.31 & 3.55 & 2.96 &  $5.71\times10^7$ & 12 \\
    \midrule
    Ours & \underline{0.2545} & \underline{0.2072} & \underline{84.78\%} &  \textbf{3.47} & \textbf{4.01} & \textbf{3.37}  & \textbf{1,024} & \textbf{3} \\
    \bottomrule
    \end{tabular}
    }
\end{table}

\subsection{Comparing with state-of-the-art Methods}
We compare the proposed method with several well-known state-of-the-art personalization methods for Stable Diffusion~\cite{stable-diffusion}, including DreamBooth~\cite{dreambooth}, Textural-Inversion~\cite{textual-inversion} and Custom Diffusion~\cite{custom-diffusion}. As shown in Figure~\ref{fig:exp_action}, given one single image as input, we evaluate the performance of several different types of generation, including the simple stylization, concept combination abilities, and two new concept interactions. Textural inversion tends to overfit the input image so most of the concepts are forgotten.  Although dreambooth~\cite{dreambooth} and custom diffusion~\cite{custom-diffusion} can successfully generate the human and the concept, the generated identities are not the same as the target image. 

\begin{figure}
  \begin{minipage}[b]{.55\textwidth}
    \centering
    \includegraphics[width=\linewidth]{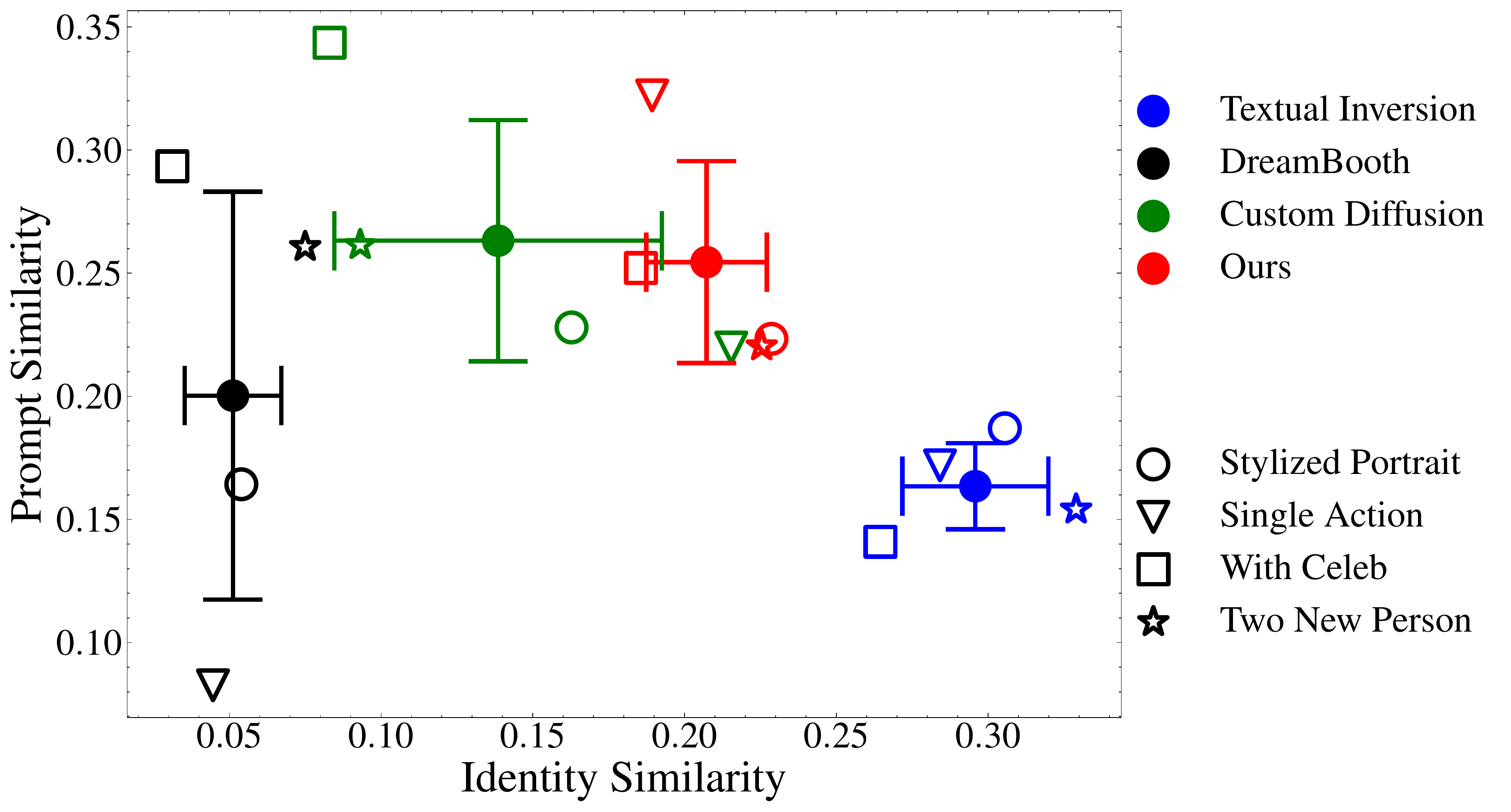}
    \caption
      {%
        Numerical analysis in terms of the prompt and identity similarity on four prompt types.
      }%
      \label{fig:analysis}%
  \end{minipage}\hfill
  \begin{minipage}[b]{.45\textwidth}
    \centering
    \resizebox{0.95\columnwidth}{!}{
    \begin{tabular}{@{}lllr@{}}
    \toprule
    Methods & Prompt$\uparrow$  & Identity$\uparrow$  & Detect$\uparrow$ \\
    \midrule
    w/o celeb basis    &   0.1386    &  \textbf{0.2528}  &   69.28\%   \\
    w/ 350 names     & 0.2214 & 0.2023 & 69.28\% \\
    w/o filter   & 0.1939 & 0.2037 & 80.62\% \\
    w flatten  & 0.2026 & 0.1873 & 80.39\% \\
    $p=64$  & \underline{0.2247} & 0.1061 & 76.47\% \\
    $p=256$   & 0.1812 & 0.0656 & 60.13\% \\
    $p=768$   & 0.1380 & 0.0836 & 47.06\% \\
    w/o $F$ & 0.1914 & 0.1896 & 55.56\% \\
    w/o aug. & 0.2083 & 0.1931 & 75.16\% \\
    \midrule
    Ours (single)      &    0.2234   &    \underline{0.2186}    &  \underline{81.05}\%   \\
    Ours (joint)  &  \textbf{0.2545}    &  0.2072  &   \textbf{84.78\%}   \\
    \bottomrule
  \end{tabular}}
    \captionof{table}
      {%
        Ablation studies.
        \label{tab:ablation}%
      }
  \end{minipage}
\end{figure}

Besides visual quality, we also perform the numerical comparison between the proposed method and baselines in Table~\ref{tab:sota}. From the table, regardless of the over-fitted textural inversion, the proposed method shows a much better performance in terms of identity similarity and the face detection rate and achieves similar text-prompt alignment as Custom Diffsuion~\cite{custom-diffusion}. We also plot the generated results on four detailed types in Figure~\ref{fig:analysis}, where the proposed method shows the best trade-off. 
Notice that, the proposed method only contains very few learning-able parameters and optimizes faster.

Moreover, since identity similarity is very subjective, we generate 200 images from different identities to form a user study. In detail, we invite 100 users to rank the generated images from one~(worst) to five~(best) in terms of visual quality, prompt alignment, and identities, getting 60k opinions in total. The results are also shown in Table~\ref{tab:sota}, where the users favor our proposed method.

\subsection{Ablation Studies}
\vspace{-0.5em}

\begin{figure}[b]
     \centering
     \vspace{-1em}
     \begin{subfigure}[b]{0.55\textwidth}
         \centering
         \includegraphics[width=\textwidth]{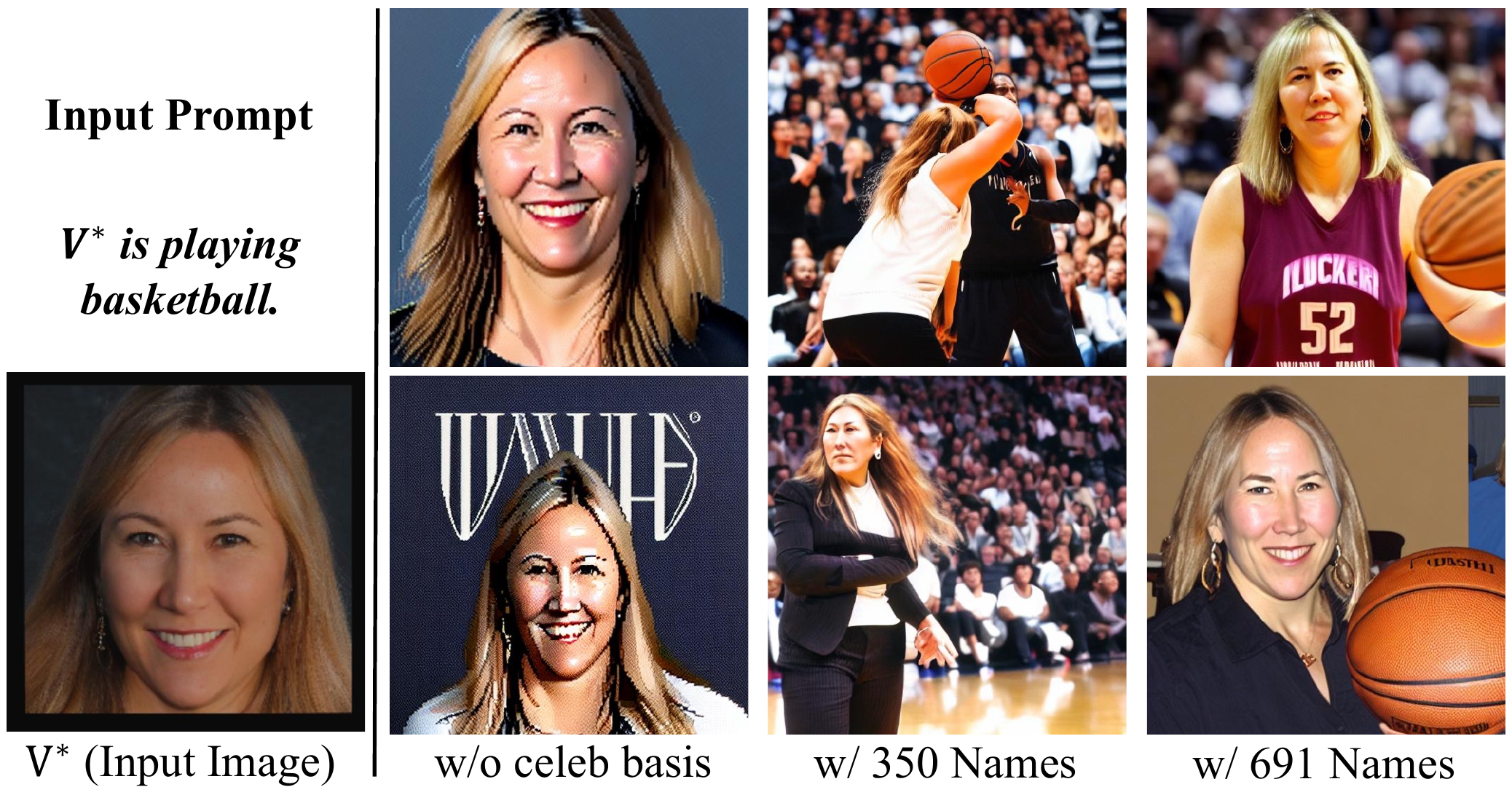}
         \vspace{-1.em}
         \caption{ \# of names in celeb basis.}
         \label{fig:ablation_celeb_basis}
     \end{subfigure}
     \hfill
     \begin{subfigure}[b]{0.42\textwidth}
         \centering
         \includegraphics[width=\textwidth]{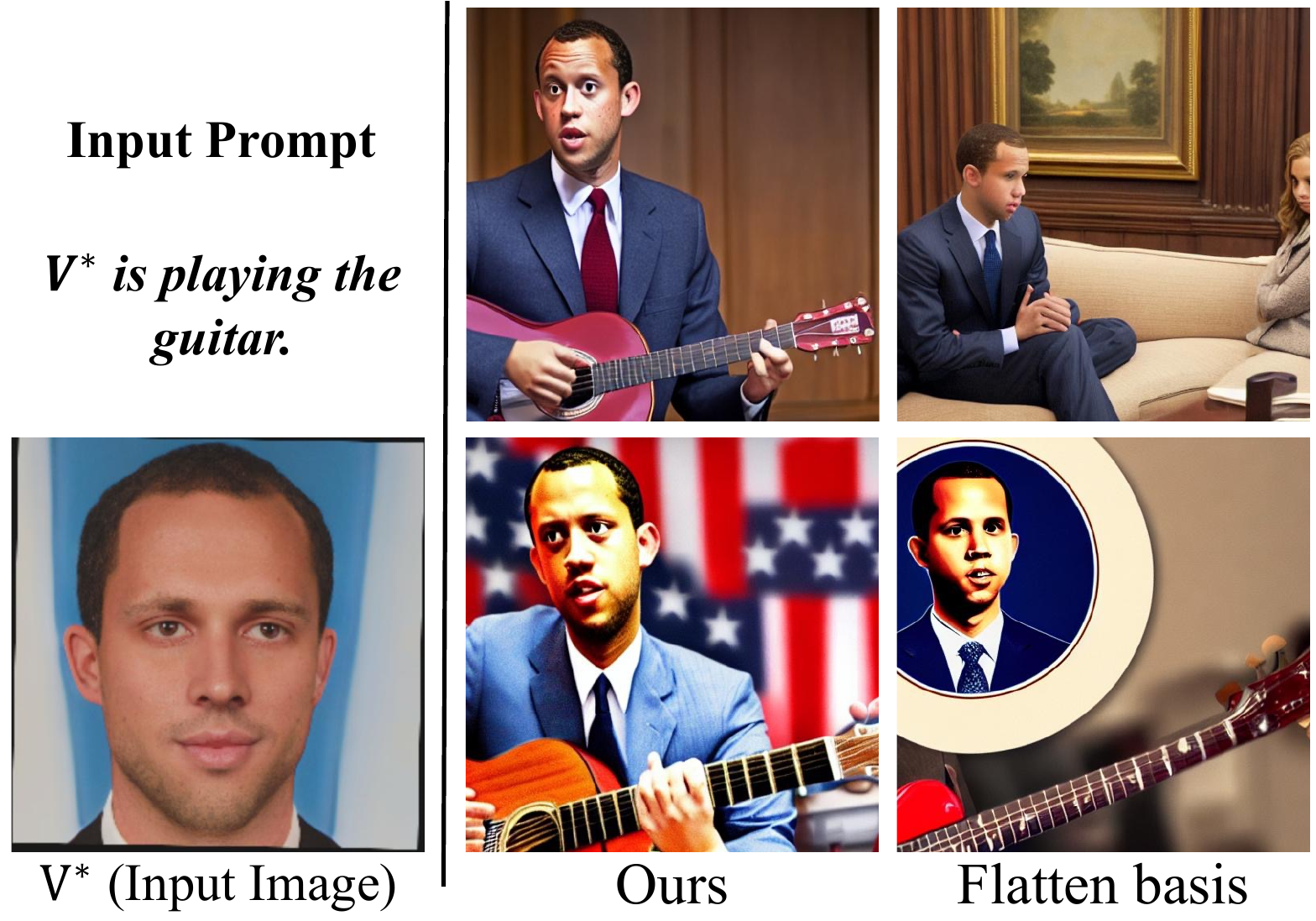}
         \vspace{-1.em}
         \caption{First and last name basis \emph{v.s.} flatten basis.}
         \label{fig:ablation_flatten_basis}
     \end{subfigure}
    \caption{Ablation studies on building celeb basis.}
        \label{fig:celeb_basis_group_abaltion}
\end{figure}

To further evaluate the sub-modules of our proposed method in both building celeb basis~(Sec.~\ref{sub:ablation_celeb_basis}) and the proposed new personalization method~(Sec.~\ref{sub:ablation_coeff_optim}), we start from the default settings ( `Ours (single)' in Table~\ref{tab:ablation}) of our method, conducting the ablation study by separately removing each submodule or using a different setting as follows. Due to the space limitation, we give more visual results in the supplementary.

\subsubsection{Ablation Studies on Celeb Basis}
\label{sub:ablation_celeb_basis}
\textbf{\# of names in celeb basis.} We evaluate the influence of the names to build a celeb basis. In extreme cases, if there is no name and we directly learn the embedding from the face encoder $F$~(w/o celeb basis), the model is overfitted to the input image and can not perform the concept combination. With fewer celeb names~(w/ 350 names), the generated quality is not good as ours baseline~(single) as in Figure~\ref{fig:ablation_celeb_basis}. Besides, the quality of the celeb basis is also important, if we do not filter the names~(w/o filter), the performance will also decrease as in Table~\ref{tab:ablation}.

\textbf{Flatten basis \emph{v.s.} first and last name basis.} In the main method, we introduce our celeb basis as the \textit{first and last name basis} since each name embedding does not have the same length. We thus involve a more naive way by flattening all the embeddings to build the basis~(w/ flatten). As shown in Fig.~\ref{fig:ablation_flatten_basis} and Table~\ref{tab:ablation}, the generated images of our first and last name basis understand prompts better.

\textbf{Choice of reduction dimension $p$.} We also evaluate the influence of the number of coefficients $p$. Considering the 768 dimensions of the CLIP text embedding, we vary $p$ ranging in $\{64,256,512,768\}$. As shown in  Table~\ref{tab:ablation}, the best result is obtained from the baseline choice~($p=512$) and we show the differences in the generated images in the supplementary materials.

\subsubsection{Ablation Studies on Coefficients Optimization}
\vspace{-0.5em}
\label{sub:ablation_coeff_optim}

\begin{figure}
\centering
\begin{subfigure}[b]{0.325\textwidth}
         \centering
         \includegraphics[width=\textwidth]{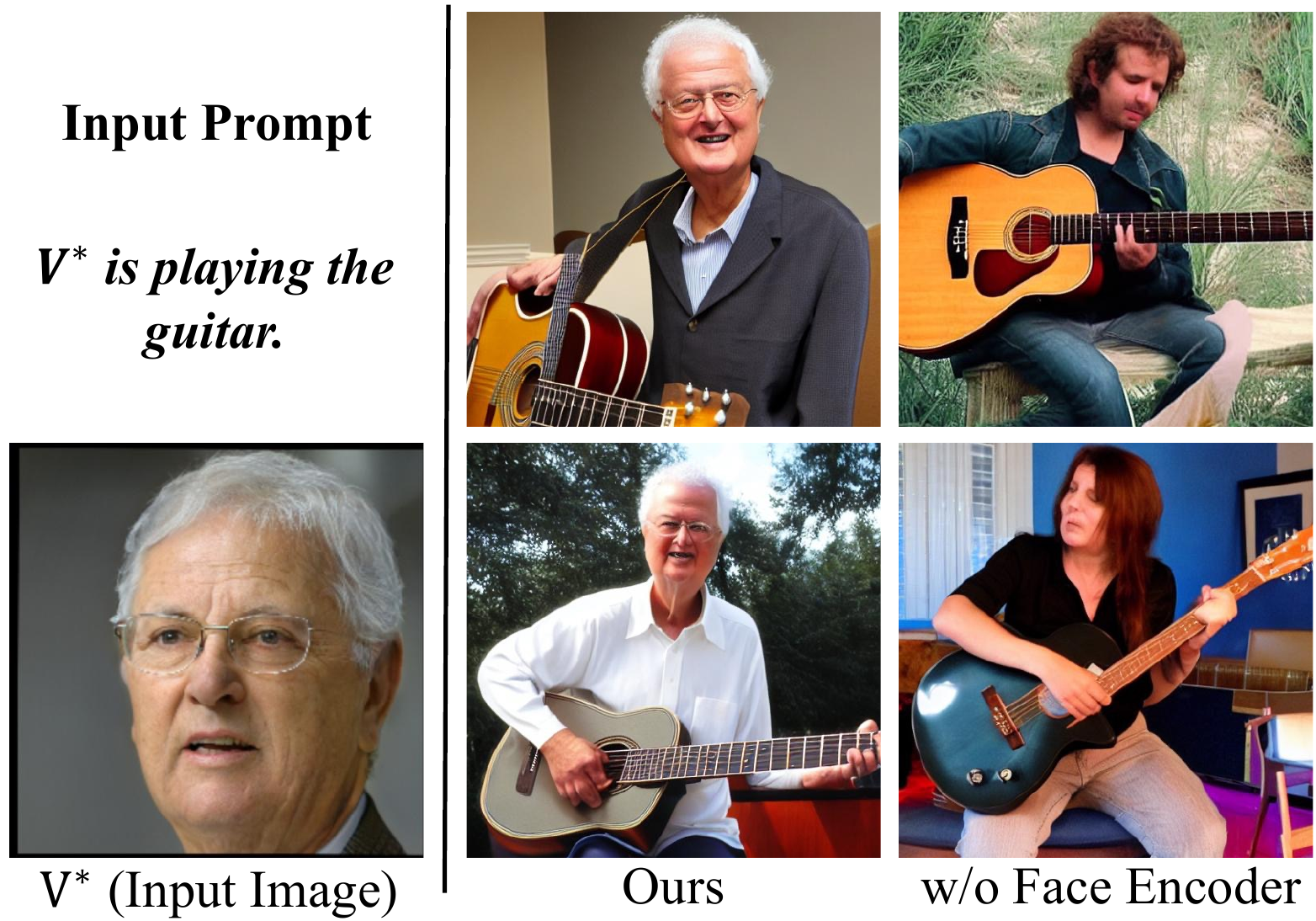}
         \vspace{-1.em}
         \caption{Face encoder.}
         \label{fig:ablation_face_encoder}
     \end{subfigure}
     \hfill
     \begin{subfigure}[b]{0.325\textwidth}
         \centering
         \includegraphics[width=\textwidth]{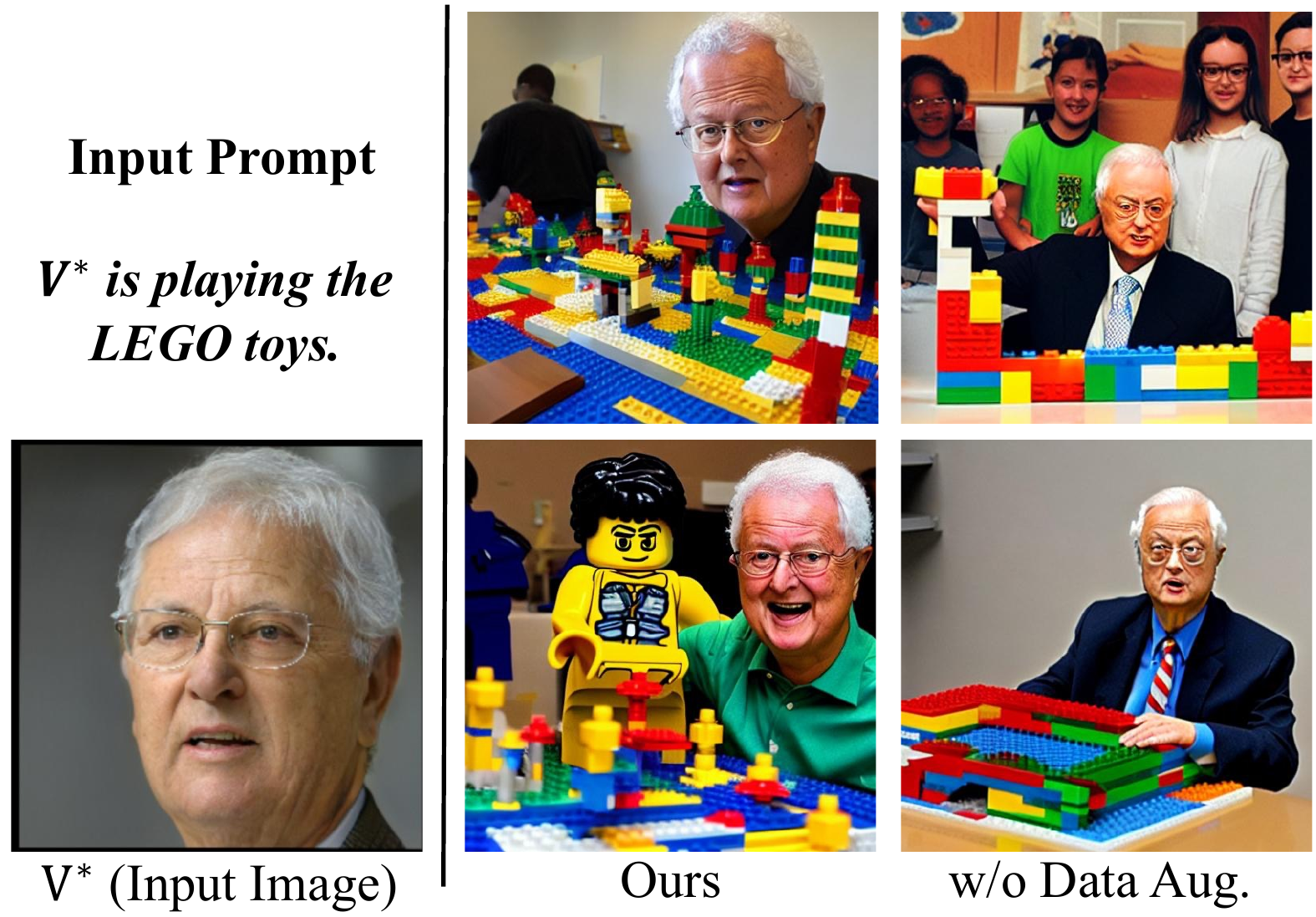}
         \vspace{-1.em}
         \caption{Data augmentation}
         \label{fig:ablation_data_aug}
     \end{subfigure}
     \hfill
     \begin{subfigure}[b]{0.325\textwidth}
         \centering
         \includegraphics[width=\textwidth]{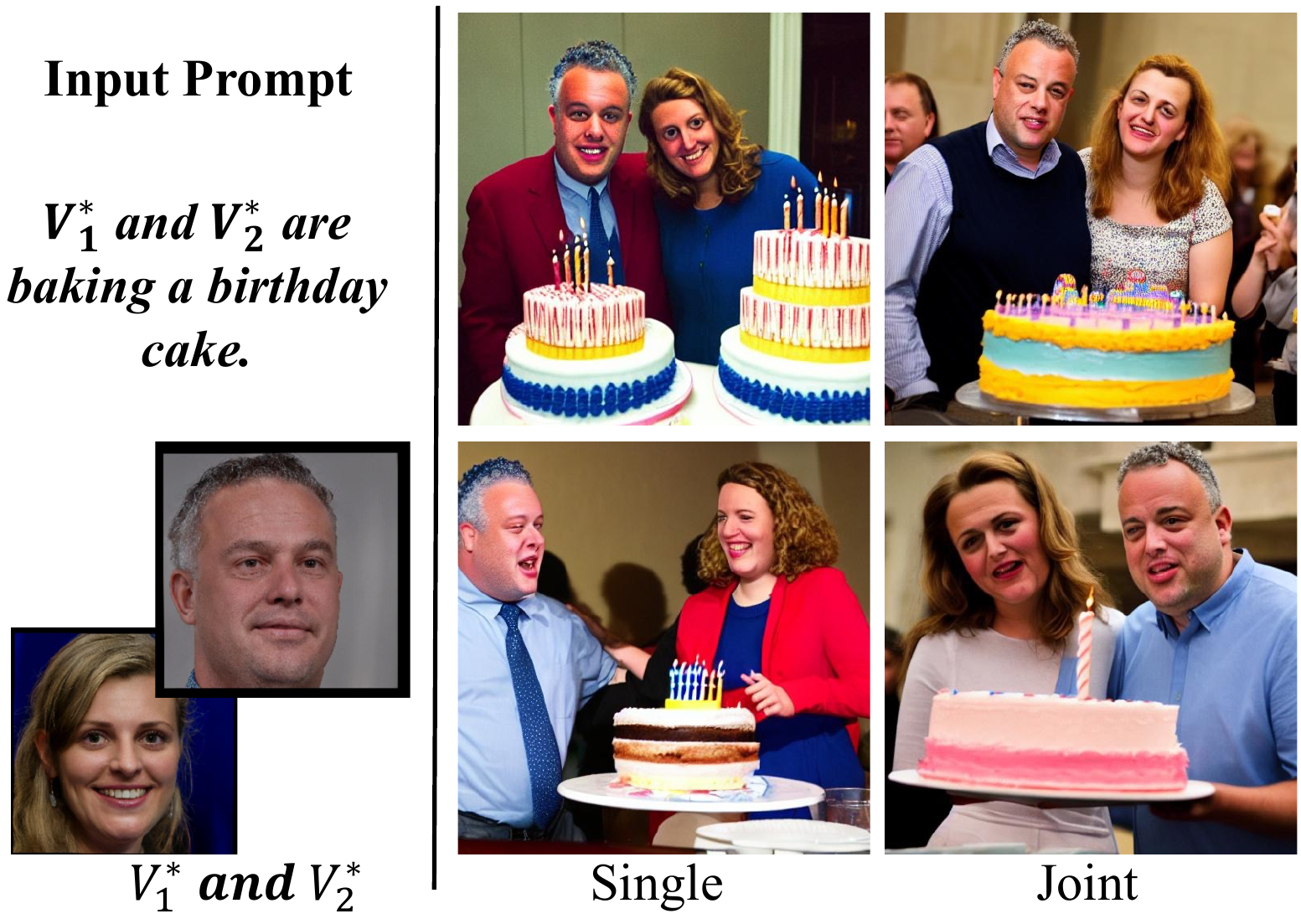}
         \vspace{-1.em}
         \caption{Single \emph{v.s.} joint optimization}
         \label{fig:ablation_joint}
     \end{subfigure}
    \caption{Ablation studies on coefficient optimization.}
    \vspace{-1.5em}
    \label{fig:ablation_all_optimization}
\end{figure}

\textbf{W/o face recognition encoder $F$.} Naively, we can optimize the coefficients $A_1, A_2$ of the celeb embeddings from back-propagation directly. However, we find the search space is still large to get satisfied results as shown in Figure~\ref{fig:ablation_face_encoder} and Table~\ref{tab:ablation}~(w/o $F$). So we seek help from the pretrained face encoder, which has more discriminative features on the face.

\textbf{W/o data augmentation.} Since there is only one single image as the tuning-able sample, we perform some data augmentations as introduced in Sec.~\ref{sec:finetune_model}. However, if we remove these augmentations, the generated face becomes vague and unnatural as shown in Figure~\ref{fig:ablation_data_aug}, and the identity of the generated samples is also decreased.

\textbf{Single \emph{v.s.} joint training.} Our method supports joint training of multiple identities using a single MLP, we evaluate the differences between single training/joint testing and joint training/joint testing. As shown in Table~\ref{tab:ablation} and Figure~\ref{fig:ablation_joint}, although training individually can also perform some interactions between the two-person, training the images jointly improve the robustness and reduces the over-fitting risk compared with single training, resulting in slightly better quantitative results.

\if
The qualitative results are shown in Figure~\ref{fig:ablation} and quantitative evaluation results are shown in Table~\ref{sample-table}.
Removing any module or using different settings lead to worse visual performance and quantitative results.
Without face encoder $F$ capturing the identity-discriminative patterns, the identity similarity drops a lot and the model takes the background as the generating target.
Without celeb basis, the model fails to perceive the input face as the `person' concept, resulting in lower prompt alignment.
Keeping the original orders of celeb embeddings like the first and last name shows more natural than mixing and flattening all celeb embeddings.
Among $\{64,256,512,768\}$, $p=512$ achieves the best performance.
Without data augmentation, the generated face becomes vague and unnatural, and the diversity decreases.
Joint training using a single MLP to fit multiple identities, which improves the robustness and reduces the over-fitting risk comparing with single training, resulting slightly better quantitative results.
\fi






\subsection{Limitation and Ethics Consideration}
\vspace{-0.5em}
\textbf{Limitation.} Although our method can successfully generate the images of the new identities, it still occurs some limitations. First, 
the real human faces of the original stable diffusion~\cite{stable-diffusion} naturally contain some artifacts, causing the naturalness of the proposed method. It might be solved by a more powerful pre-trained text-to-image model~(\eg, Imagen~\cite{imagen}, IF~\cite{if}) since they can generate better facial details. Secondly, we only focus on human beings currently. It is also interesting to build the basis of other species, \eg, cars, and cats, we leave it as future work.

\textbf{Ethics Consideration and Broader Impacts.} We propose a new method for model personalization on the human face, which might be used as a tool for deepfake generation since we can not only generate the human face but also can interact with other people. However, this prevalent issue is not limited to this approach alone, as it also exists in other generative models and content manipulation techniques. Besides, our personalization person generation method can also ablate the abilities of the erasing concept methods~\cite{gandikota2023erasing} and other deep fake detection methods~\cite{Luo_2021_CVPR}.

\section{Conclusion}
\label{sec:conclusion}
We propose a new method to personalize the pre-trained text-to-image model on a  specific kind of concept, \ie, the human being, with simply single-shot tuning. Our approach is enabled by defining a basis in the domain of the known celebrity names' embeddings. Then, we can map the facial feature from the pre-trained face recognition encoder to reconstruct the coefficients of the new identity. Compared with the previous concept injection method, our method shows stronger concept combination abilities, \eg, better identity preservation, can be trained on various identities at once, and can perform some interacting abilities between the newly-added humans. Besides, the proposed method only requires 1024 parameters for each person and can be optimized in under 3 minutes, which is also much more efficient than previous methods.

{
\bibliographystyle{unsrt}
\bibliography{11_references}
}

\appendix
\label{sec:appendix}

\section{More Qualitative Results}



\subsection{Evaluation on Faces in a Wider Range} 
Besides evaluating the results on StyleGAN synthetic faces as in the main paper, we compare our method with state-of-the-art methods on a wide range of human face images, including the real word faces, the faces from different races, and the interactions of the same gender.
These face images are from VGGFace2~\cite{cao2018vggface2} and collected from the Web.

\begin{figure*}[h]
    \centering
    \includegraphics[width=\textwidth]{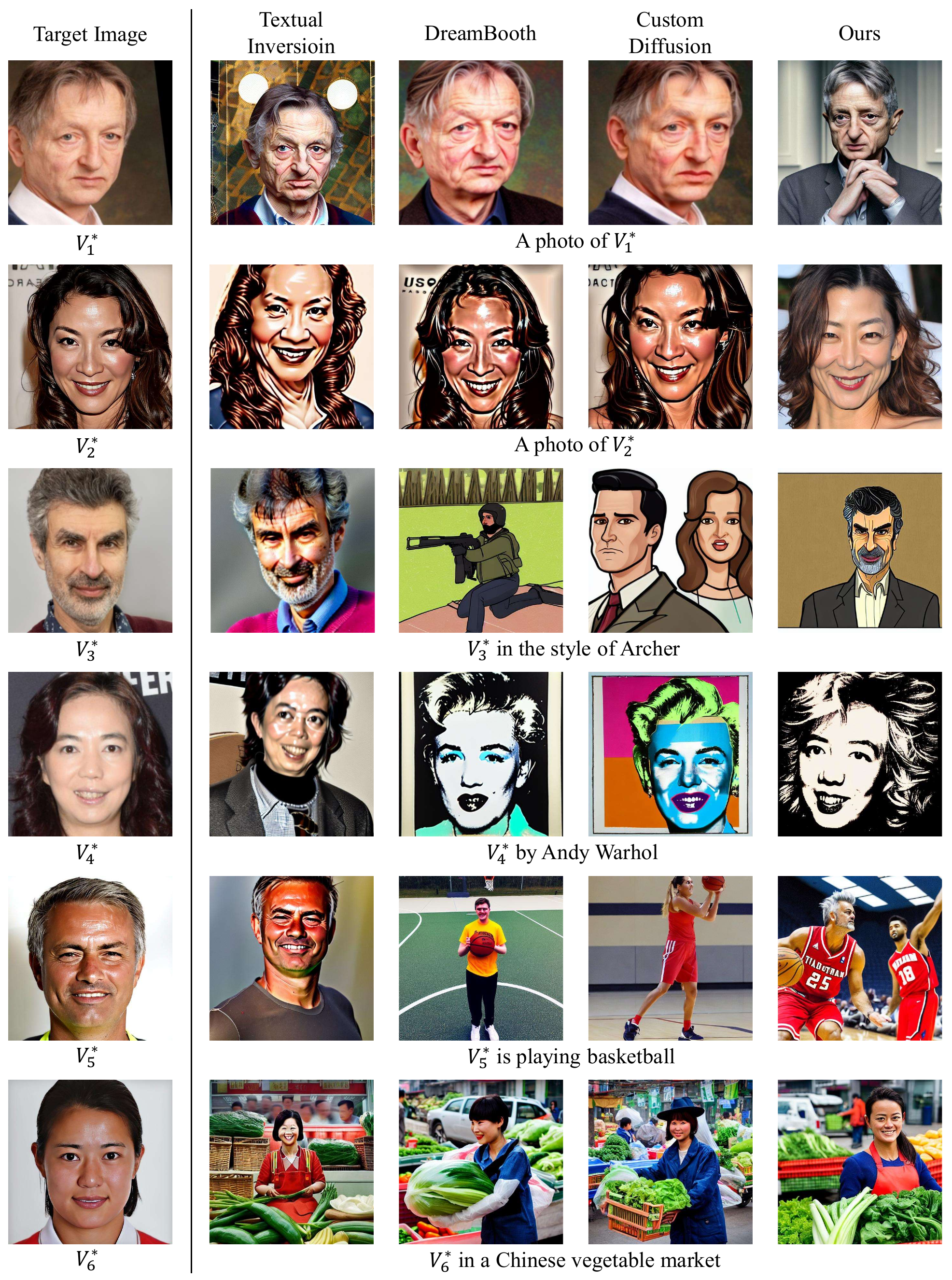}
    \caption{Single person's personalization result for \textbf{real} identities.}
    \label{fig:supp_single}
\end{figure*}

\textbf{Single Person in an Image.} Figure~\ref{fig:supp_single} shows the evaluation results on a single person's personalization for real identities.
Concretely, Row 1-2 shows the ability to generate diverse images of different methods.
The baselines Textual Inversion~\cite{textual-inversion}, DreamBooth~\cite{dreambooth}, and Custom Diffusion~\cite{custom-diffusion} can generate identity-consistent results with the single prompt `A photo of $V^*$', but the light condition and skin texture seem not natural.
DreamBooth and Custom Diffusion fail to disentangle the background of the target inputs.
The stylization results are shown in Row 3-4, where Textual Inversion preserves the target identities but fails to change the style, DreamBooth, and Custom Diffusion generate the images of the corresponding style but lose the identity information, and our method preserves the identity and completes the stylization as given prompts.

\begin{figure*}[t]
    \centering
    \includegraphics[width=\textwidth]{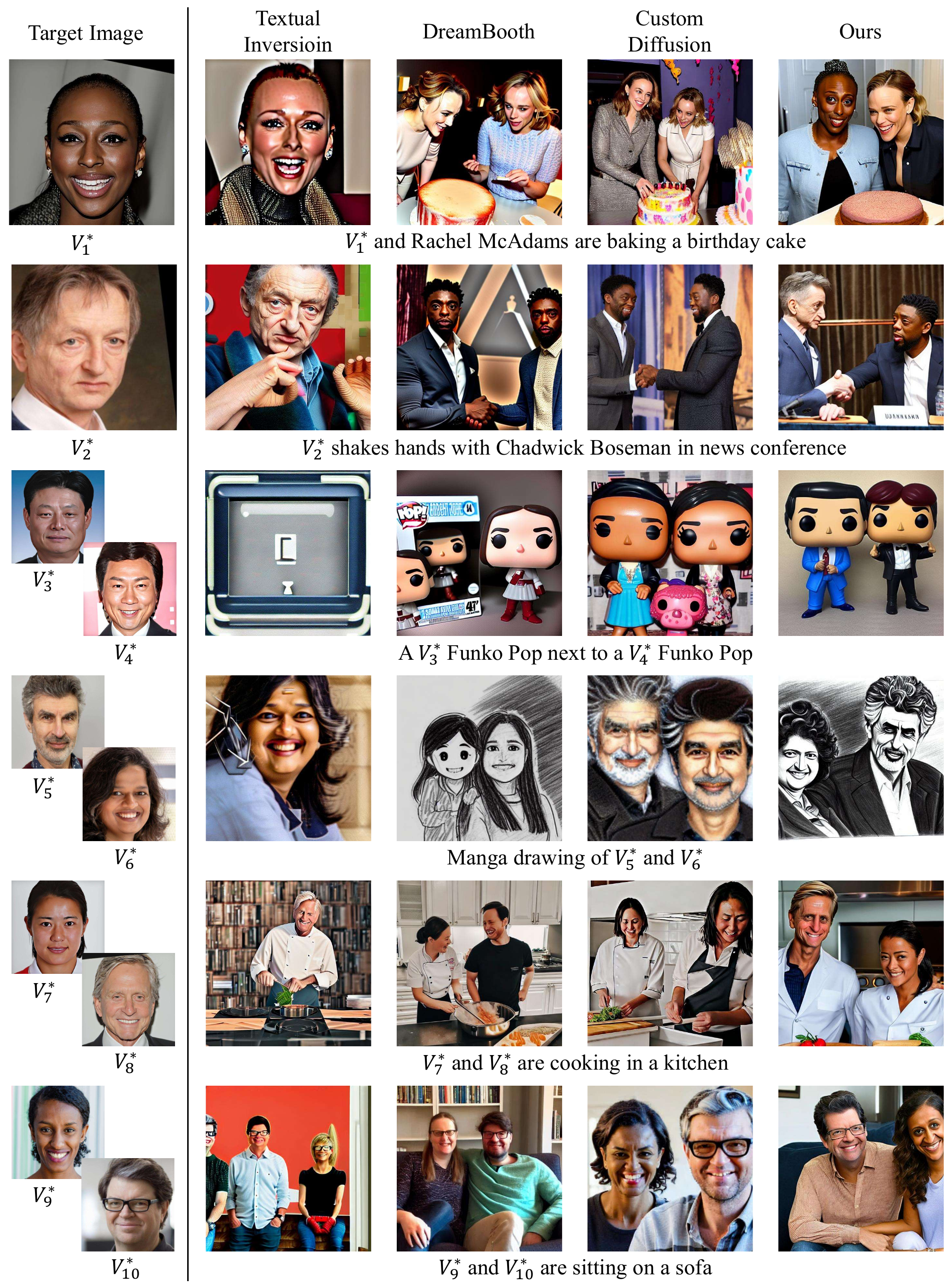}
    \caption{Multiple persons' personalization result for \textbf{real} identities.}
    \label{fig:supp_two}
\end{figure*}

\textbf{Multiple Persons in an Image.} We give the comparison of generating multiple-person interaction in Figure~\ref{fig:supp_two}. Similar to the single-person results, the proposed method shows a good concept combination ability among different persons~(including the newly added and the original celebrities) and can serve as a novel concept to communicate with other humans.

\subsection{Additional Results Comparing with concurrently work: FastComposer~\cite{xiao2023fastcomposer}}

Different from the methods~\cite{textual-inversion, dreambooth, custom-diffusion} that the finetuning time cost for each identity is below 30 minutes, FastComposer~\cite{xiao2023fastcomposer} needs to be pre-trained on a large human dataset, requiring 150k steps with batch size 128 on 8 NVIDIA A6000~(48GB) GPUs. So as its performance is restricted by the dataset as in their limitation. 
Differently, our method requires only 400 steps with batch size 2 on a single NVIDIA A100 (40GB) GPU. After the pre-training stage, FastComposer can generate images based on the given single image and prompt like other personalized methods.


To further demonstrate the efficacy of our method, we evaluate the qualitative performance of our method compared with FastComposer.
Our method outperforms FastComposer on single action controlling, the interaction between two persons, and expression controlling.
The detailed comparisons are as follows.

\begin{figure}[h]
    \centering
    \includegraphics[width=\textwidth]{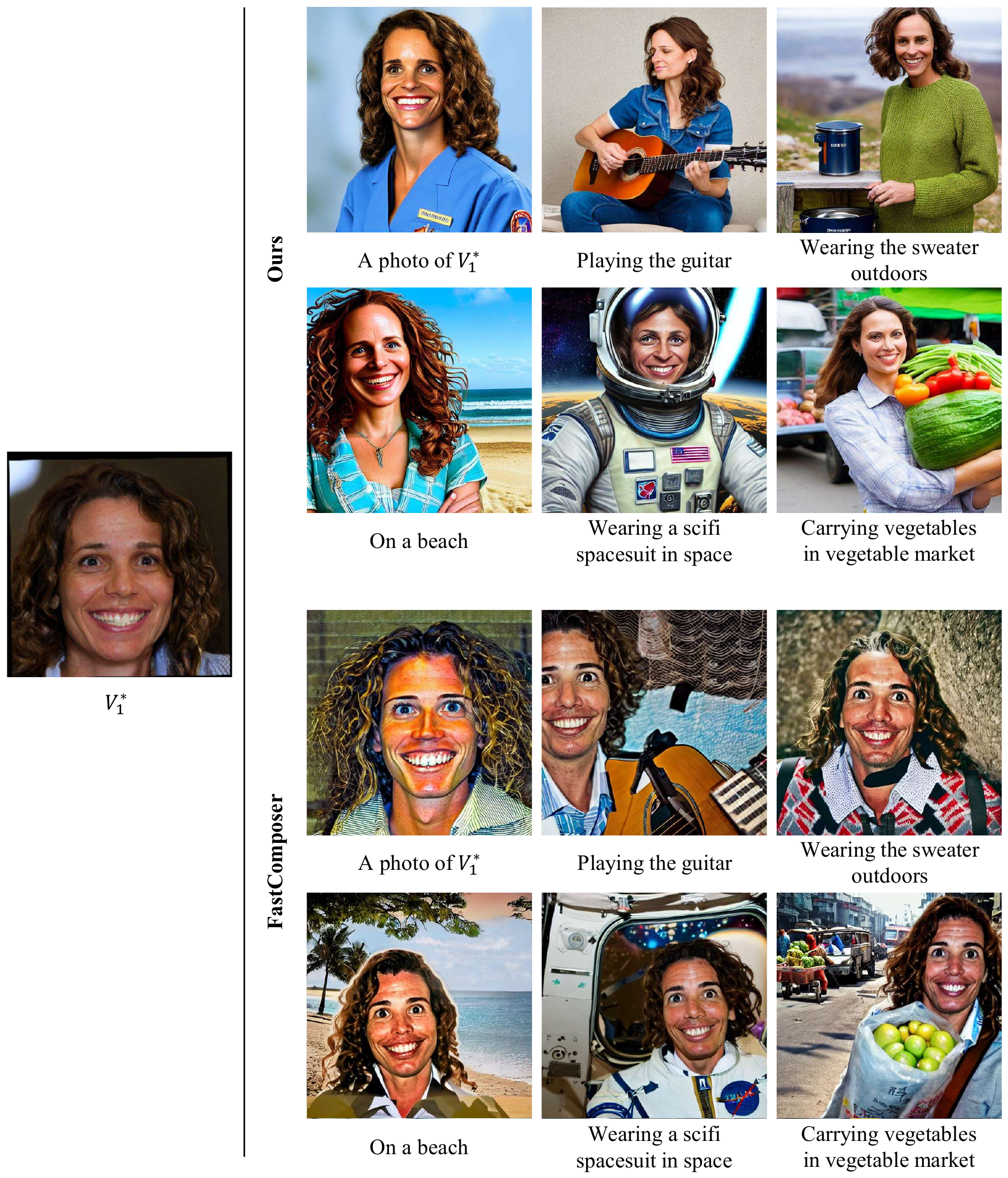}
    \caption{Additional single person's personalization results comparing with FastComposer~\cite{xiao2023fastcomposer}.}
    \label{fig:supp_stylegan_single_woman}
\end{figure}
\begin{figure}[h]
    \centering
    \includegraphics[width=\textwidth]{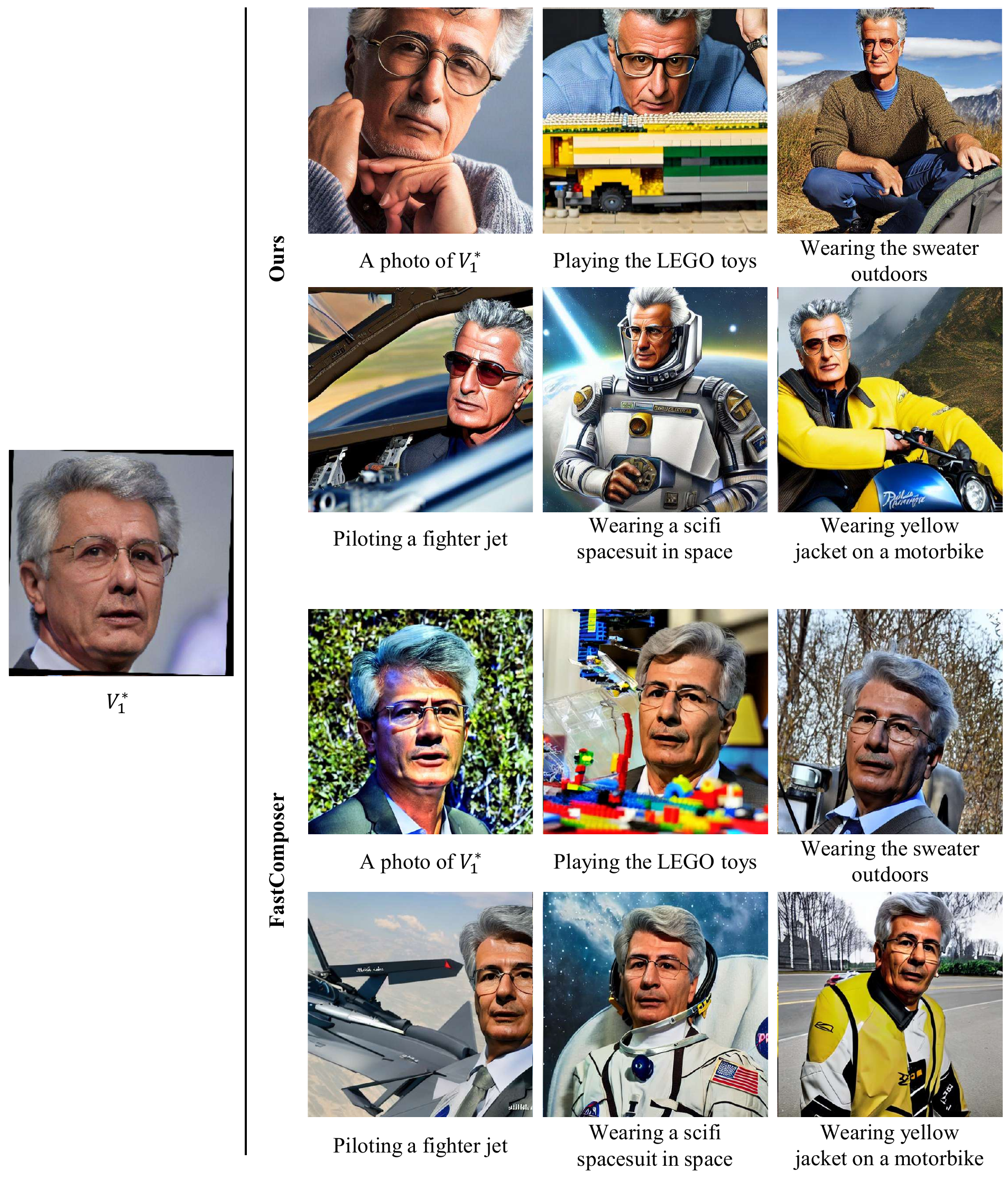}
    \caption{Additional single person's personalization results comparing with FastComposer~\cite{xiao2023fastcomposer}.}
    \label{fig:supp_stylegan_single_man}
\end{figure}

\textbf{Action of Single Person.}
We first evaluate the performance of single persons' personalization.
As illustrated in Figure~\ref{fig:supp_stylegan_single_woman} and Figure~\ref{fig:supp_stylegan_single_man}, we show the synthesized results under six scenarios in each figure. 

In Figure~\ref{fig:supp_stylegan_single_woman}, given the prompt `A photo of $V^*_1$', FastComposer generates an unnatural result.
In the remaining scenarios, the light seems very disharmonious in the images synthesized by FastComposer.
Besides, in these results, only the upper chest and head can be generated, where the human body and limbs fail to appear, revealing an over-fitting issue of FastComposer.

In Figure~\ref{fig:supp_stylegan_single_man}, FastComposer meets the same issue.
Although the key objects and face identity are almost consistent with the input prompts, e.g. LEGO toys, fighter jet, and yellow jacket, the results from FastComposer seem to be a rude combination of the objects and face.
Besides, their methods have a huge issue with concept forget. Some important prompts like `sweater' and `motorbike' are ignored by FastComposer.

We also highlight the faces generated by FastComposer share very similar expressions from the original input image, i.e. dilated pupils, wrinkles on the forehead, and mouth shape, which means this method is overfitted on the input image. Differently, our method generates identity- and prompt-consistent results, appearing natural and realistic.

\begin{figure}[h]
    \centering
    \includegraphics[width=\textwidth]{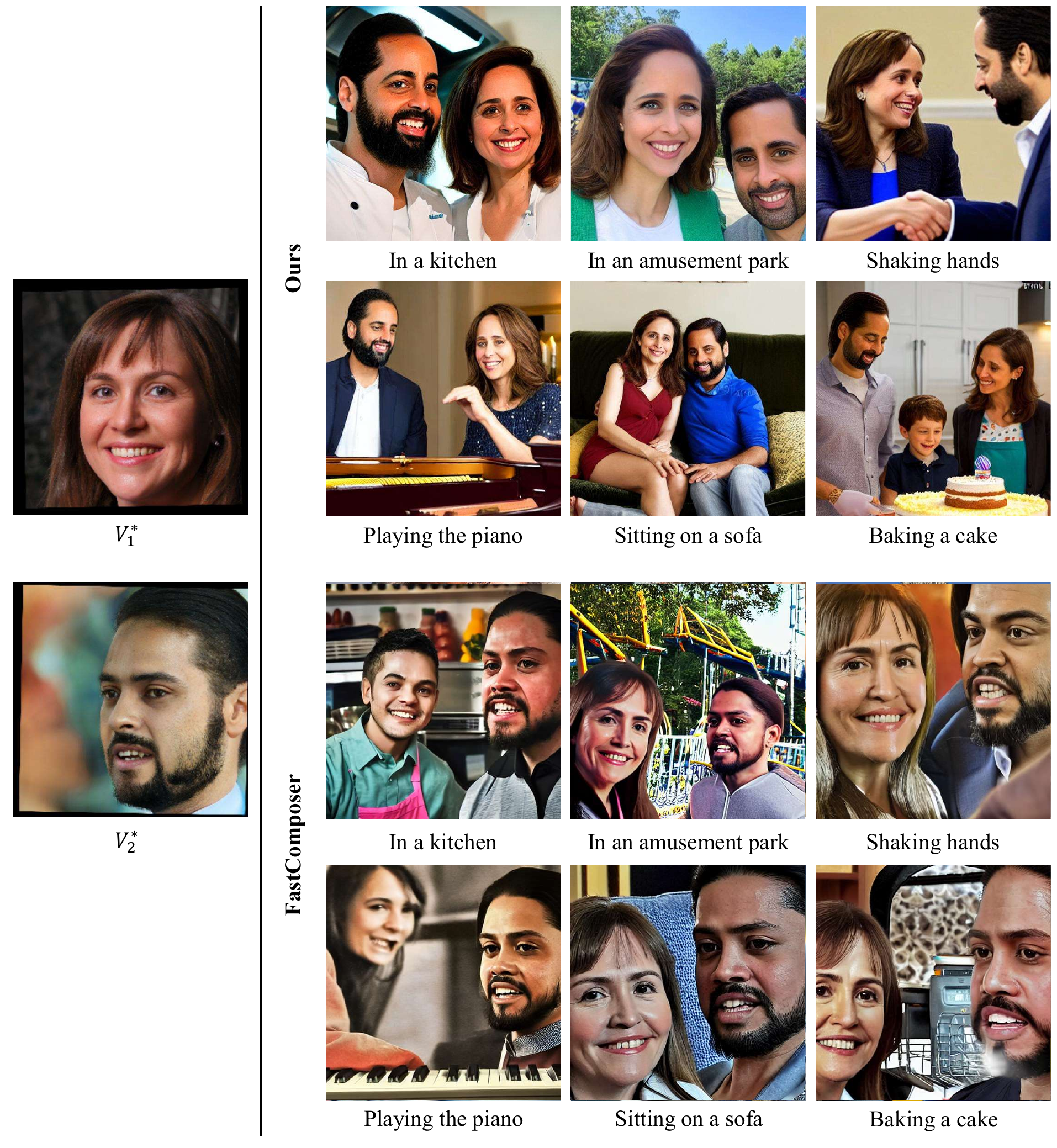}
    \caption{Additional multiple persons' interaction personalization results comparing with FastComposer~\cite{xiao2023fastcomposer}.}
    \label{fig:supp_stylegan_two}
\end{figure}

\textbf{Interaction between Two Persons.}
Figure~\ref{fig:supp_stylegan_two} illustrates the interaction and common action of two persons in a single image.
Sharing the similar aforementioned over-fitting problem, FastComposer only generates the face part of the human.
As for the actions including `shaking', `playing', `sitting', and `baking', FastComposer fails to generate the correct behaviors, where the faces usually occupy the most part of the images. The problem of invariant expression also occurs in FastComposer.

\begin{figure}[h]
    \centering
    \includegraphics[width=\textwidth]{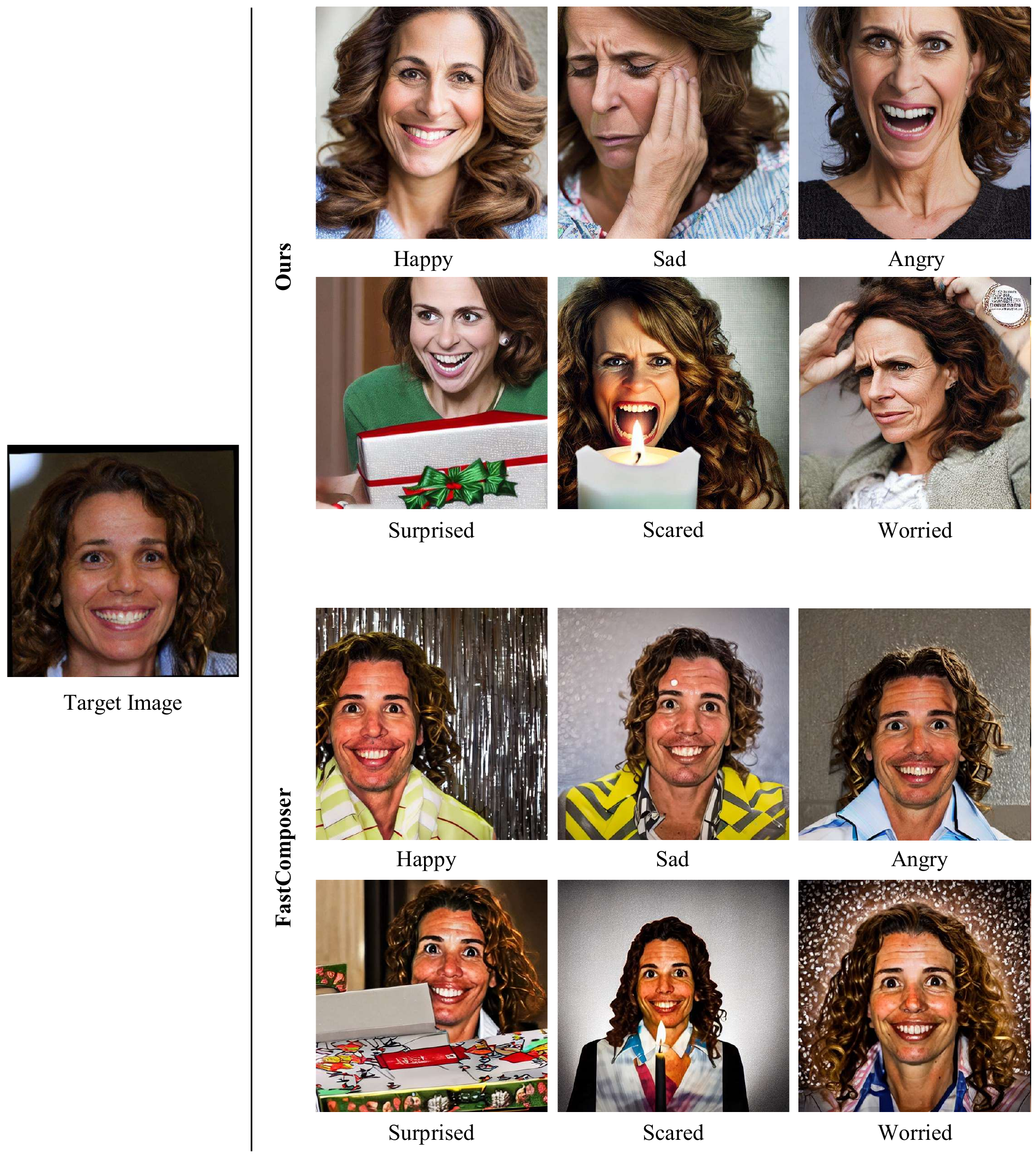}
    \caption{Comparing with FastComposer~\cite{xiao2023fastcomposer}, our method has a better ability of controlling face expression.}
    \label{fig:supp_expression}
\end{figure}

\textbf{Expression Controlling.}
\label{sec:expression_controlling}
Considering that the above FastComposer results struggle to generate the different expressions, we conduct a face expression controlling experiment to validate the 
abilities of our method and FastComposer.  We use the following prompts to change the expression of the target input image:
\begin{itemize}
    \item $V^*$ is smiling with happiness
    \item $V^*$ is crying with sadness
    \item $V^*$ has an anger expression
    \item $V^*$ looks very surprised, looking at a gift box
    \item $V^*$ is shaking with fear scared by a candle lighting, scared expression
    \item $V^*$ has a tired expression, headache, uncomfortable, a frown creased her forehead
\end{itemize}
As shown in Figure~\ref{fig:supp_expression}, the proposed method can successfully generate different expressions under the text prompts, which shows the advantage of the proposed method.

\section{Implementation Details}


\subsection{Celeb Images Generated by Stable Diffusion}

\begin{figure}[h]
    \centering
    \includegraphics[width=\textwidth]{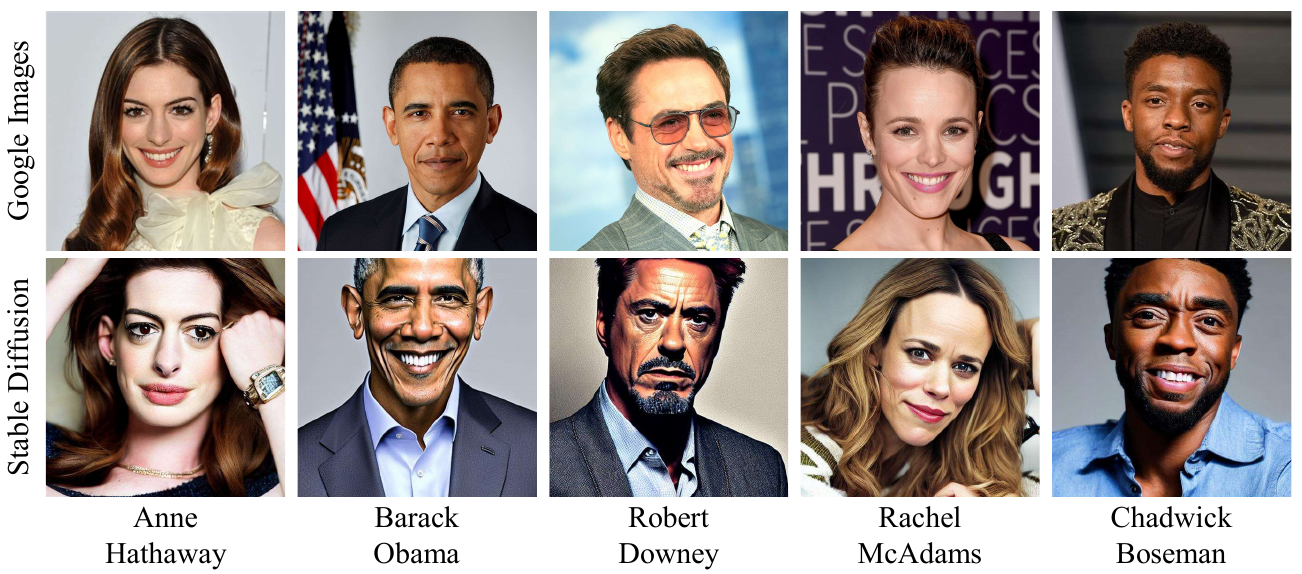}
    \caption{Stable Diffusion has the ability to generate identity-consistent celeb images. But in the aspect of facial details, the celeb images synthesized by Stable Diffusion may exaggerate some facial features compared with the real ones collected from Google Images.}
    \label{fig:supp_celeb}
\end{figure}

In general, the celeb images generated by Stable Diffusion~\cite{stable-diffusion} and those realistic ones collected from Google Images seem very similar, but with a little difference on facial details.
As seen in Figure~\ref{fig:supp_celeb}, we compare the real celeb images collected from Google Images with the fake ones synthesized by Stable Diffusion.
Overall, the real and fake images closely resemble the same individual. However, certain facial features in the fake images may be exaggerated, which can constrain our method's performance and explain why the celebrity profiles in the outcomes generated by both baseline models and our approach do not precisely match their real counterparts.

\subsection{Collecting and Filtering Celeb Names}
\label{sec:supp_filter}

\begin{figure}[t]
    \vspace{-2em}
    \centering
    \includegraphics[width=\textwidth]{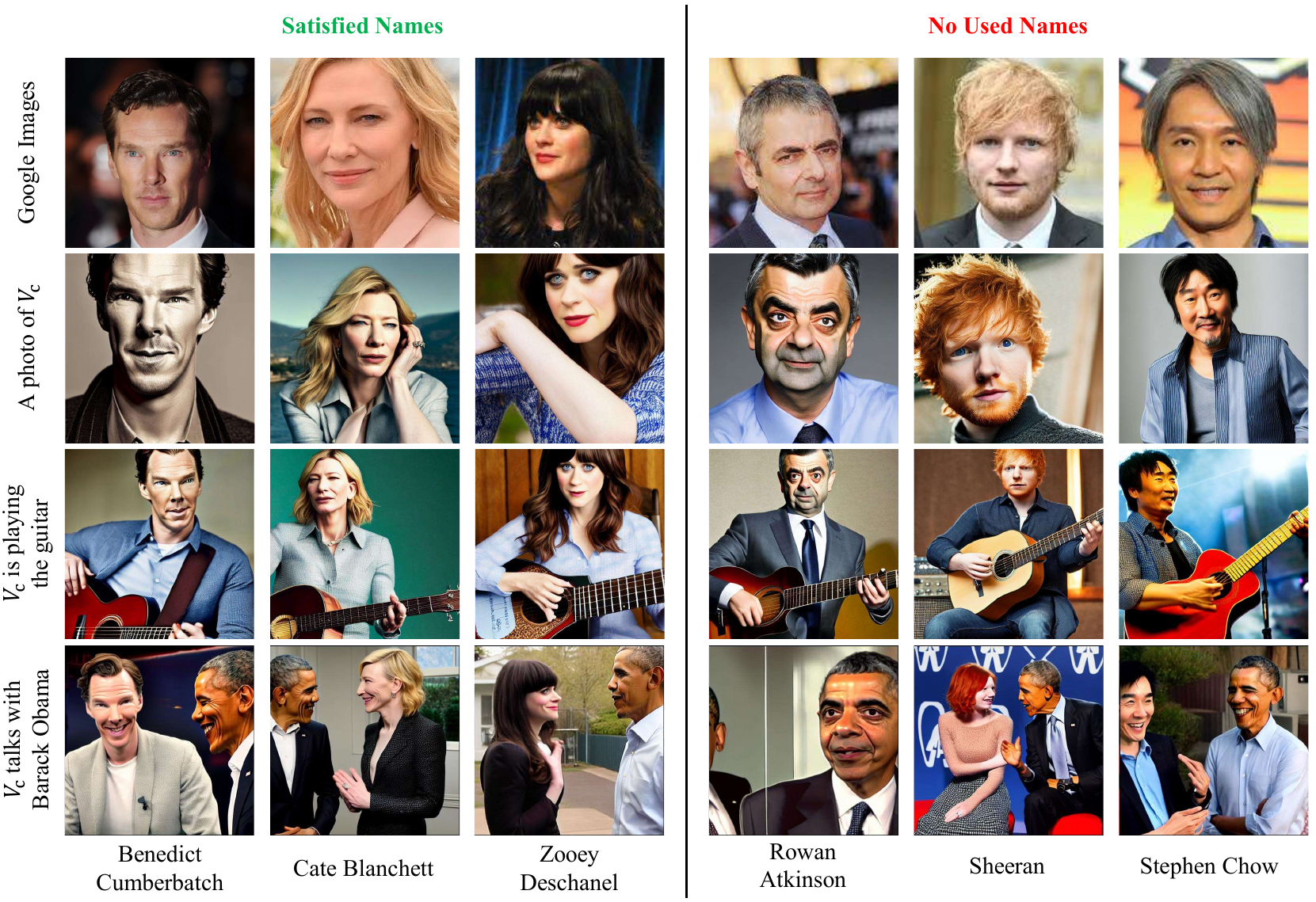}
    \vspace{-1em}
    \caption{We filter out the celeb names that can generate prompt- and identity-consistent images through Stable Diffusion~\cite{stable-diffusion}. Left: the satisfied names that can generate images share the same identity with Google Images results. Furthermore, these names have the ability to interact with objects or other satisfied celebs (e.g. `Barack Obama'). Right: Stable Diffusion confuses the face of `Rowan Atkinson' and `Barack Obama' in row four. The gender of `Sheeran' is mistaken as a girl in row four. The identity of `Stephen Chow' has a large gap with that of Google Images results.}
    \label{fig:supp_implementation_filter_name}
\end{figure}

After crawling about 1,500 celebrity names, to filter out the names that have the ability to generate prompt-consistent identities and interacting with other celebs, we feed three types of prompts, i.e. `A photo of $V_{\rm{c}}$', `$V_{\rm{c}}$ is playing the guitar', and `$V_{\rm{c}}$ talks with Barack Obama', to the Stable Diffusion~\cite{stable-diffusion} for synthesizing image results, where $V_{\rm{c}}$ indicates the celeb name (e.g. `Anne Hathaway').
Figure~\ref{fig:supp_implementation_filter_name} shows the examples that satisfy the filtering condition and the ones that fail to generate reasonable results.
Consequently, 691 names pass the checking, which can be tokenized and encoded into $m=691$ celeb embedding groups.

\subsection{Building Celeb Basis based on PCA}

We only keep the first and second embeddings of each celeb embedding group, resulting in the first name embedding set $C_1$ and the second name embedding set $C_2$, where $C_1, C_2\in\mathbb{R}^{m\times d}$.
Then, for simplicity, we omit the subscript of $C_k$, using $C$ to indicate any one of $C_1$ and $C_2$.

\textbf{Calculation of Mean.}
Considering $C$ may have repetitive embeddings (each row corresponds an embedding vector), for each $C$, we first remove the duplicate rows that come from the same token to make sure each token only occurs once at most.
Then we calculate the mean $\overline{C}\in\mathbb{R}^d$ across the second dimension of $C$ as mentioned in the main paper.

\textbf{PCA.}
The PCA algorithm has many different coding implementations in practice.
In our method, we use Singular Value Decomposition (SVD) to skip the calculation of covariance matrix in PCA.
Please refer to~\cite{pca, pca_python} for more detailed theoretical demonstration and coding techniques.
Due to the built celeb basis is not optimized, the PCA process only needs to be computed once during the training stage.

\subsection{Training Recipe}
We train the MLP with a learning rate of 0.005 and batch size of 2 on a single NVIDIA A100 GPU.
The training augmentation includes horizontal flip, color jitter, and random scaling ranging in $0.1\sim 1.0$.
For single identity training, the optimization costs 400 steps, taking $\sim3$ minutes.
For 10 identities joint training, we found that training 2,500 steps is enough, taking $\sim18$ minutes (averaged about 250 steps and 108 seconds for each identity).
The text prompts for training include:

\begin{itemize}
    \item A photo of a face of $V^*$ person
    \item A rendering of a face of $V^*$ person
    \item The photo of a face of $V^*$ person
    \item A rendition of a face of $V^*$ person
    \item A illustration of a face of $V^*$ person
    \item A depiction of a face of $V^*$ person
\end{itemize}



\section{More Ablation Studies}

\begin{figure}[h]
     \centering
     \begin{subfigure}[b]{0.6\textwidth}
         \centering
         \includegraphics[width=\textwidth]{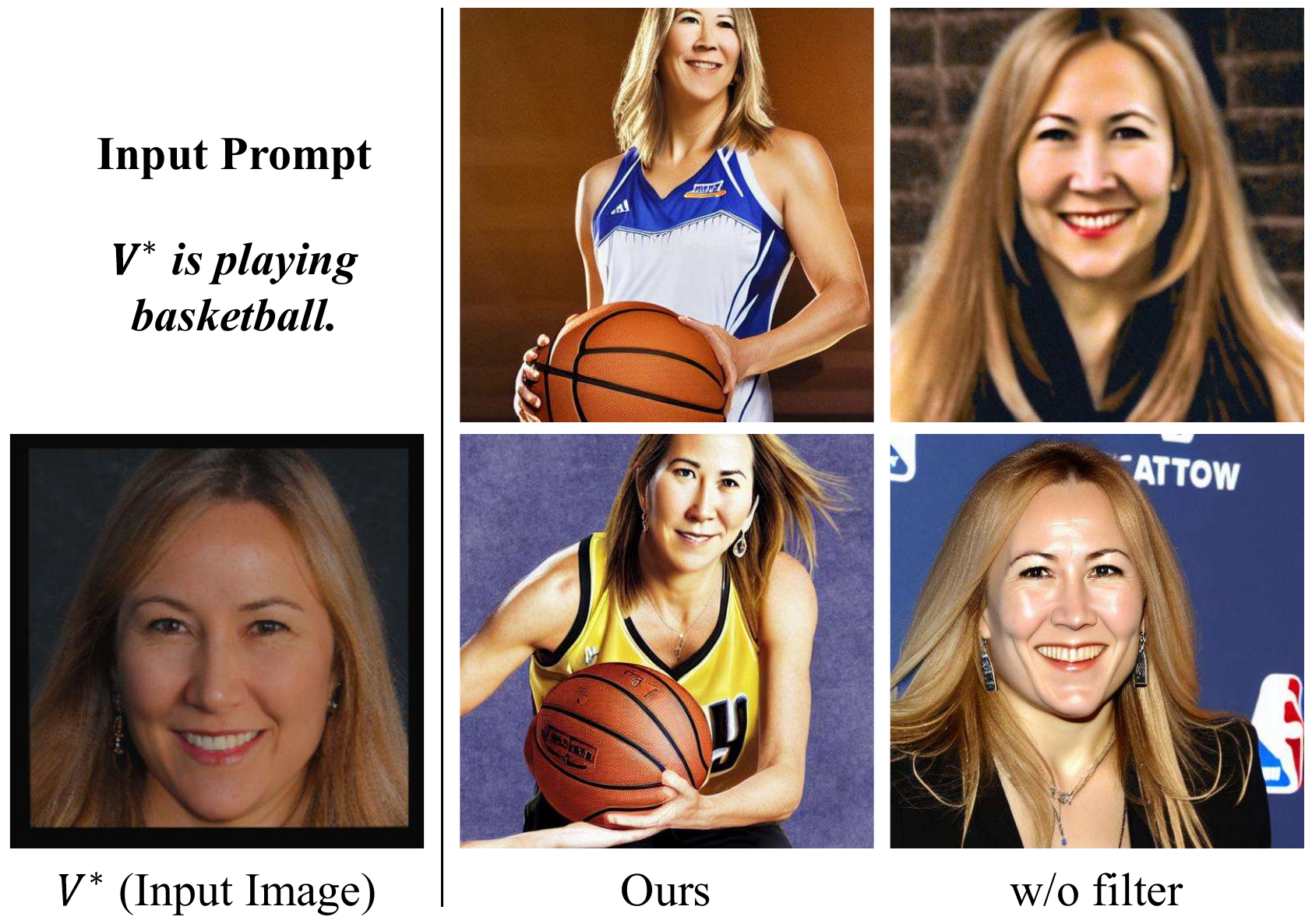}
         \caption{The influence of filtering the useless celeb names or not.}
         \label{fig:supp_ablation_filter}
     \end{subfigure}
     \hfill
     \begin{subfigure}[b]{0.95\textwidth}
         \centering
         \includegraphics[width=\textwidth]{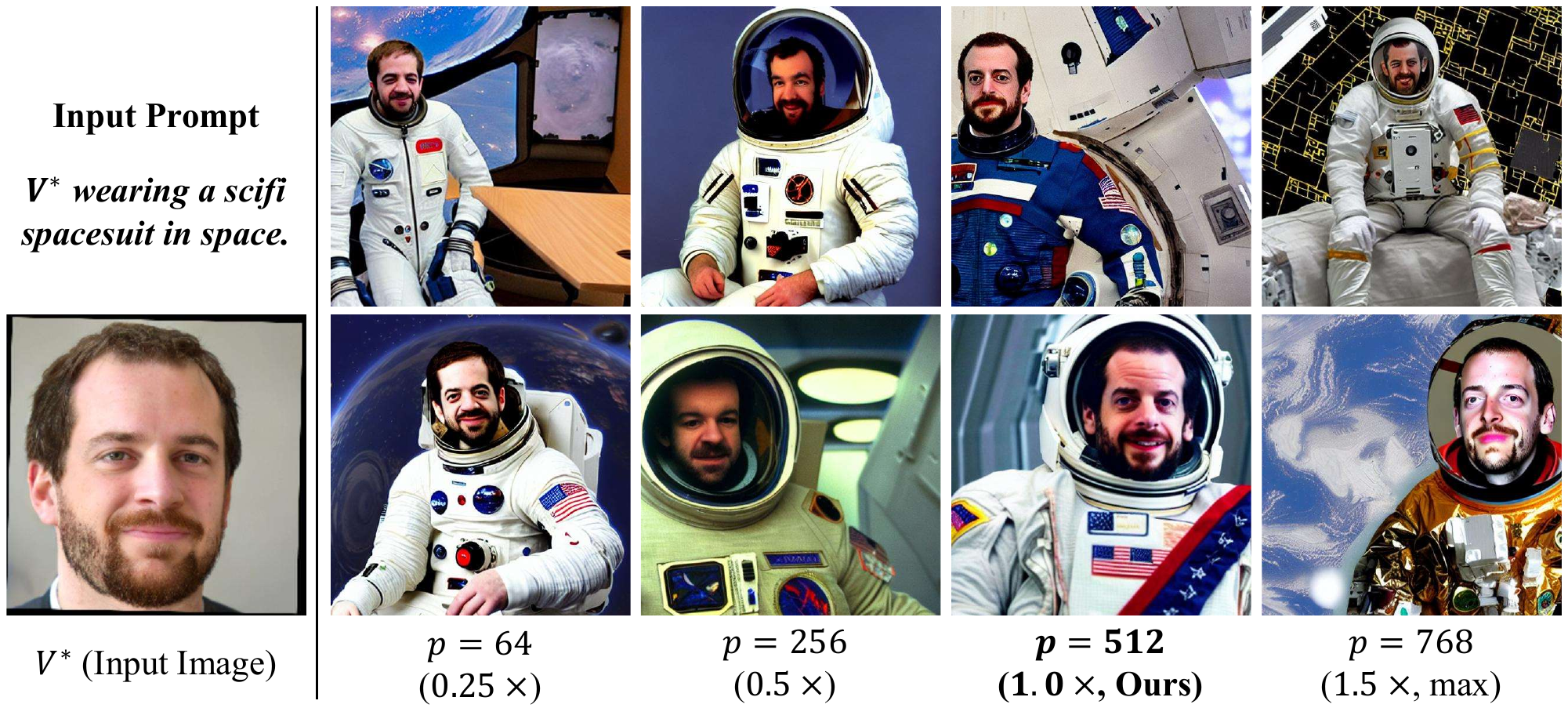}
         \caption{The ablation on the values of $p$.}
         \label{fig:supp_ablation_p}
     \end{subfigure}
    \caption{Additional ablation studies on building celeb basis and the choice of reduction dimension $p$.}
        \label{fig:supp_ablation}
\end{figure}



\textbf{W/o filtering celebrity names.}
As mentioned in Section~\ref{sec:supp_filter}, Stable Diffusion fails to generate correct images from some celeb names.
In our method, we manually drop these bad samples.
If this filtering process canceled, the interaction ability of the learned identities drops a lot, as shown in Figure~\ref{fig:supp_ablation_filter}.

\textbf{Variants of reduction dimension $p$.}
In our method, the PCA reduction dimension $p$ $(p<k)$ controls the degrees of freedom for modulating the variance applied to the celeb basis mean.
Considering that $k=768$ in the CLIP~\cite{clip} text encoder, we conduct a series experiments to study the choice of $p$.
Four values of $p$ are chosen, including $64, 256, 512, 768$.
Note that $p=768$ means PCA is not used.
Larger $p$ costs more memory storage.
As shown in Figure~\ref{fig:supp_ablation_p}, chosing $p=512$ comes to the best identity similarity, which is consistent with the quantitative results in our main paper.





\end{document}